\documentclass{article} 
\usepackage{iclr2024_conference,times}

\usepackage{amsmath,amsfonts,bm}

\def\eqref#1{equation~\ref{#1}}

\def\1{\bm{1}}

\DeclareMathAlphabet{\mathsfit}{\encodingdefault}{\sfdefault}{m}{sl}
\SetMathAlphabet{\mathsfit}{bold}{\encodingdefault}{\sfdefault}{bx}{n}

\DeclareMathOperator*{\argmax}{arg\,max}

\usepackage{hyperref}
\usepackage{url}
\usepackage{inconsolata}
\usepackage{amssymb}
\usepackage{amsmath}
\usepackage{xspace}
\usepackage{graphicx}
\usepackage{booktabs}
\usepackage{multirow}
\usepackage{caption}
\usepackage{subcaption}
\usepackage[normalem]{ulem} 
\usepackage{todonotes}
\usepackage{enumitem}
\usepackage{colortbl}
\newcommand{\ours}{DoLa\xspace}

\oddsidemargin=\dimexpr\oddsidemargin + 1cm\relax
\evensidemargin=\dimexpr\evensidemargin + 1cm\relax
\marginparwidth=\dimexpr \marginparwidth + 2cm\relax
\usepackage{array}
\usepackage{geometry}

\usepackage{tikz}
\usepackage{pgfplots}
\usepgfplotslibrary{groupplots}
\pgfplotsset{compat=1.16}

\title{DoLa: Decoding by Contrasting Layers Improves Factuality in Large Language Models}

\newcommand\blfootnote[1]{%
  \begingroup
  \renewcommand\thefootnote{}\footnote{#1}%
  \addtocounter{footnote}{-1}%
  \endgroup
}

\author{Yung-Sung Chuang$^\dagger{}^\star$, Yujia Xie$^\ddagger$, Hongyin Luo$^\dagger$, Yoon Kim$^\dagger$, James Glass$^\dagger$, Pengcheng He$^\ddagger$\\
$^\dagger$Massachusetts Institute of Technology, $^\ddagger$Microsoft\\
\texttt{yungsung@mit.edu, yujiaxie@microsoft.com}\\
\texttt{\{hyluo,yoonkim,glass\}@mit.edu, herbert.he@gmail.com}\\
}

\iclrfinalcopy 
\begin{document}

\maketitle

\begin{abstract}
Despite their impressive capabilities, large language models (LLMs) are prone to hallucinations, i.e., generating content that deviates from facts seen during pretraining. We propose a simple decoding strategy for reducing hallucinations with pretrained LLMs that does not require conditioning on retrieved external knowledge nor additional fine-tuning. Our approach obtains the next-token distribution by contrasting the differences in logits obtained from projecting the later layers versus earlier layers to the vocabulary space, exploiting the fact that factual knowledge in an LLMs has generally been shown to be localized to particular transformer layers. We find that this \textbf{D}ecoding by C\textbf{o}ntrasting \textbf{La}yers (\ours) approach is able to better surface factual knowledge and reduce the generation of incorrect facts. 
 \ours consistently improves the truthfulness across multiple choices tasks and open-ended generation tasks, for example improving the performance of LLaMA family models on TruthfulQA by 12-17\% absolute points, demonstrating its potential in making LLMs reliably generate truthful facts.\footnote{The source code is available at \url{https://github.com/voidism/DoLa}.}
\blfootnote{$^\star$Work mainly done during an internship at Microsoft.}
\end{abstract}

\vspace{-15pt}
\section{Introduction}
\vspace{-5pt}
Large language models (LLMs) have demonstrated great potential in numerous natural language processing (NLP) applications~\citep{gpt3, chatgpt2023, openai2023gpt4}. However, despite the continued increase in performance and the emergence of new capabilities  from scaling LLMs \citep{wei2022emergent}, their tendency to ``hallucinate'', i.e., generate content that deviates from real-world facts observed during pretraining~\citep{ji2023survey}, remains a persistent challenge. This represents a major bottleneck in their deployment especially for high-stakes applications (e.g., clinical/legal settings) where reliable generation of trustworthy text is crucial.

While the exact reasons for LMs' hallucinations are not fully understood, a possible reason  is due to the maximum likelihood language modeling objective which minimize the forward KL divergence between the data and model distributions. This objective potentially results in a model with mass-seeking behavior which causes the LM to assign non-zero probability to sentences that are not fully consistent with knowledge embedded in the training data. Empirically, an LM trained with the next-word prediction objective on finite data has been shown to result in a model that uses linguistic knowledge to recognize the superficial patterns, instead of recognizing and generating the real-world facts extracted from the training corpus~\citep{ji2023survey}. 

From a model interpretability perspective, transformer LMs have been loosely shown to encode  ``lower-level''  information  (e.g., part-of-speech tags) in the earlier layers, and more ``semantic'' information in the later layers~\citep{tenney2019bert}. More recently, \citet{dai2022knowledge} find that ``knowledge neurons'' are distributed in the topmost layers of the pretrained BERT model. \citet{meng2022locating} show that factual knowledge can even be edited by manipulating a specific set of feedforward layers within an autoregressive LM. We propose to exploit this modular encoding of knowledge to amplify the factual knowledge in an LM through a {contrastive decoding} approach, where the output next-word probability is obtained from the \emph{difference} in logits between a higher layer versus a lower layer. By emphasizing the knowledge of higher layers and downplaying that of lower layers, we can potentially make LMs more factual and thus reduce hallucinations. 

\begin{figure}[t!]
\begin{center}
\includegraphics[width=0.6\textwidth]{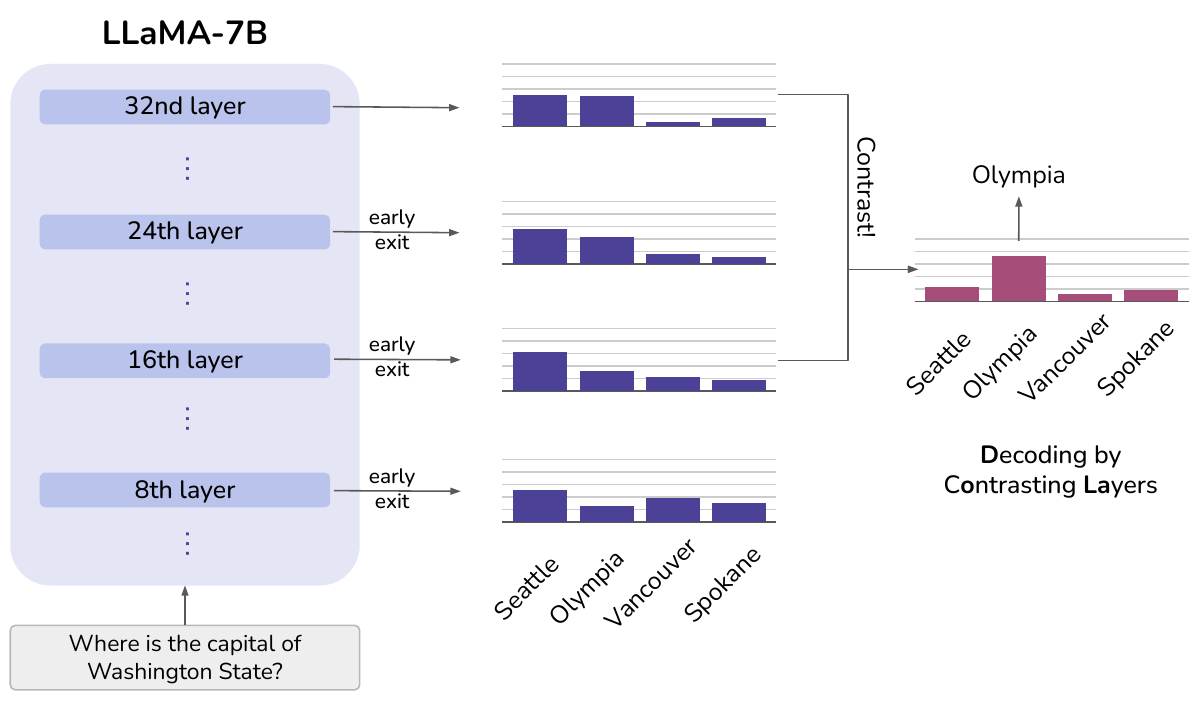}
\end{center}
\vspace{-10pt}
\caption{Illustration of an LLM progressively incorporates factual information along layers. While the next-word probabilities of ``\textit{Seattle}'' remain similar throughout different layers, the probabilities of the correct answer ``\textit{Olympia}'' gradually increase from lower to higher layers. \ours uses this fact to decode by contrasting the difference between layers to sharpen an LLM's probability towards factually correct outputs.}
\vspace{-10pt}
\label{fig:dola}
\end{figure}

An illustration of this idea for a simple example is shown in Figure~\ref{fig:dola}. While ``\textit{Seattle}'' maintains high probability throughout all the layers---presumably because it is a syntactically plausible answer---the probability of the true answer ``\textit{Olympia}'' increases after the higher layers inject more factual knowledge. Contrasting the differences between the different layers can thus reveal the true answer in this case. Based on this concept, we propose a new decoding method, \textbf{D}ecoding by C\textbf{o}ntrasting \textbf{La}yers (DoLa), for better surfacing factual knowledge embedded in an LLM without retrieving external knowledge or additional fine-tuning. 

Experiments on TruthfulQA~\citep{lin2022truthfulqa} and FACTOR~\cite{muhlgay2023generating} demonstrate that \ours is able to increase the truthfulness of the models of the LLaMA family~\citep{touvron2023llama}.  Further experiments on chain-of-thought reasoning for StrategyQA~\citep{geva2021did} and GSM8K~\citep{cobbe2021training} also show that it can facilitate more factual reasoning. Finally, experiments using GPT-4 for open-ended chatbot evaluation~\citep{vicuna2023} show that when compared with the original decoding method, \ours can generate informative and significantly more factual responses that lead to better ratings from GPT-4. From an efficiency perspective, we find that \ours causes only a small additional latency in the decoding process, suggesting it as a practical and useful decoding strategy for improving the truthfulness of LLMs.

\vspace{-5pt}
\section{Method}
\label{sec:method}
\vspace{-7pt}
Recent language models consist of an embedding layer, $N$ stacked transformer layers, and an affine layer 
 $\phi(\cdot)$ for predicting the next-word distribtution. Given a sequence of tokens $\{x_1, x_2, \dots, x_{t-1}\}$, the embedding layer first embeds the tokens into a sequence of vectors $H_0=\{h_1^{(0)}, \dots, h_{t-1}^{(0)}\}$. Then $H_0$ would be processed by each of the transformer layers successively. We denote the output of the $j$-th layer as $H_j$. Then, the vocabulary head $\phi(\cdot)$  predicts the probability of the next token $x_{t}$ over the vocabulary set $\mathcal{X}$,
\vspace{-3pt}
\begin{align*}
    p(x_{t} \mid x_{<t}) = \mathrm{softmax}\bigl(\phi(h_t^{(N)})\bigr)_{x_{t}}, \quad x_{t}\in\mathcal{X}.
\end{align*}
Instead of applying $\phi$ on the final layer, our approach contrasts the higher-layer and lower-layer information to obtain the next-token probability. More specifically, for the $j$-th early layer, we also compute the next-token probability using $\phi(\cdot)$ as follows, where $\mathcal{J} \subset \{0, \dots, N-1\}$ is a set of candidate layers,
\begin{align*}
    q_j(x_{t} \mid x_{<t}) = \mathrm{softmax}\bigl(\phi(h_t^{(j)})\bigr)_{x_{t}}, \quad j\in {\mathcal{J}}.
\end{align*}
The idea of applying language heads directly to the hidden states of the middle layers, known as \textit{early exit}~\citep{teerapittayanon2016branchynet,elbayad2020depth,schuster2022confident}, has proven to be effective even without special training process~\citep{kao2020bert}, as the residual connections~\citep{he2016deep} in transformer layers make the hidden representations gradually evolve without abrupt changes. 
Using $q_j(x_{t})$ to represent $q_j(x_{t} \mid x_{<t})$ for notational brevity, we then compute the probability of the next token by,
\vspace{-2pt}
\begin{align*}
    \hat{p}(x_{t} \mid x_{<t}) & = \mathrm{softmax}\bigl(\mathcal{F}\bigl(q_N(x_{t}), q_M(x_{t})\bigr)\bigr)_{x_t}, \\
    \text{where} \quad M & = \argmax_{j\in\mathcal{J}} \,\, d\big(q_N(\cdot), q_j(\cdot)\bigr). \nonumber
\end{align*}
Here, layer $M$ is named \textit{premature layer}, while the final layer, i.e., layer $N$, is named \textit{mature layer}. 
The operator $\mathcal{F}(\cdot, \cdot)$, to be elaborated further in Section~\ref{sec:contrast}, is used to contrast between the output distributions from the premature layer and the mature layer by computing the log-domain difference between two distributions.
The premature layer is dynamically selected in each decoding step using a distributional distance measure $d(\cdot, \cdot)$ (we use Jensen-Shannon Divergence) between the mature layer and all the candidate layers in $\mathcal{J}$. We discuss $d(\cdot, \cdot)$ in more detail in  Section~\ref{sec:dynamic}. 
The motivation for selecting the layer with the highest distance $d(\cdot, \cdot)$ is to ensure that the model would significantly change its output after that selected layer, and thus have a higher chance to include more factual knowledge that does not exist in the early layers before it.

\begin{figure}[t!]
\begin{center}
\includegraphics[width=0.85\textwidth]{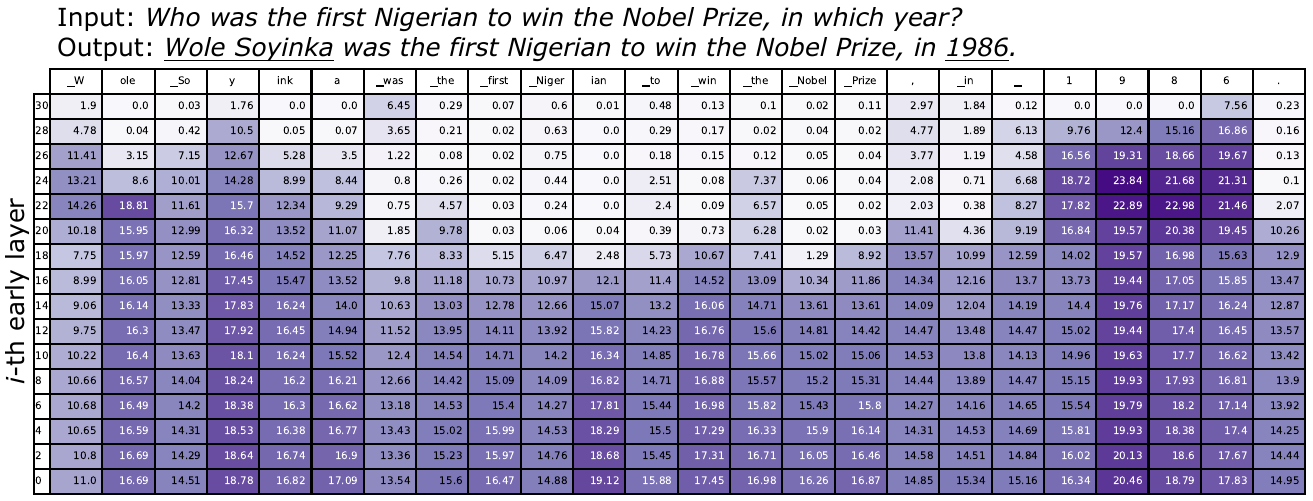}
\end{center}
\vspace{-10pt}
\caption{JSD (scaled by $10^{5}$) between the final 32nd layer and even-numbered early layers. Column names are decoded tokens in each step. Row names are indices of the early layers. 0 means word embedding layer.}
\label{fig:jsdiv}
\vspace{-20pt}
\end{figure}
\vspace{-8pt}
\subsection{Factual Knowledge Evolves Across Layers}
\label{sec:locate}
\vspace{-6pt}

We conduct preliminary analysis with 32-layer LLaMA-7B~\citep{touvron2023llama} to motivate our approach. We compute the Jensen-Shannon Divergence (JSD) between the early exiting output distributions $q_j( \cdot \mid x_{<t})$ and the final layer output distribution $q_N( \cdot \mid x_{<t})$, to show how the early exiting outputs are different from the final layer outputs. 
Figure~\ref{fig:jsdiv} shows the JSDs when decoding the answer for the input question, from which we can observe two patterns.
\textbf{Pattern \#1} happens when predicting important name entities or dates, such as \emph{Wole Soyinka} and \emph{1986} in Figure~\ref{fig:jsdiv}, which require factual knowledge. We observe the calculated JSD would be still extremely high in the higher layers. This pattern indicates that the model is still changing its predictions in the last few layers, and potentially injecting more factual knowledge into the predictions.
\textbf{Pattern \#2} happens when predicting function words, such as \emph{was, the, to, in}, and the tokens copied from the input question, such as \emph{first Nigerian, Nobel Prize}. When predicting these ``easy'' tokens, we can observe that the JSD becomes very small from middle layers. This finding indicates that the model has already decided what token to generate in middle layers, and keeps the output distributions almost unchanged in the higher layers. This finding is also consistent with the assumptions in early exiting LMs~\citep{schuster2022confident}. A preliminary analysis that can quantitatively support this observation is also shown in Appendix~\ref{appx:ner}.

\begin{figure}[t!]
\begin{center}
\includegraphics[width=0.7\textwidth]{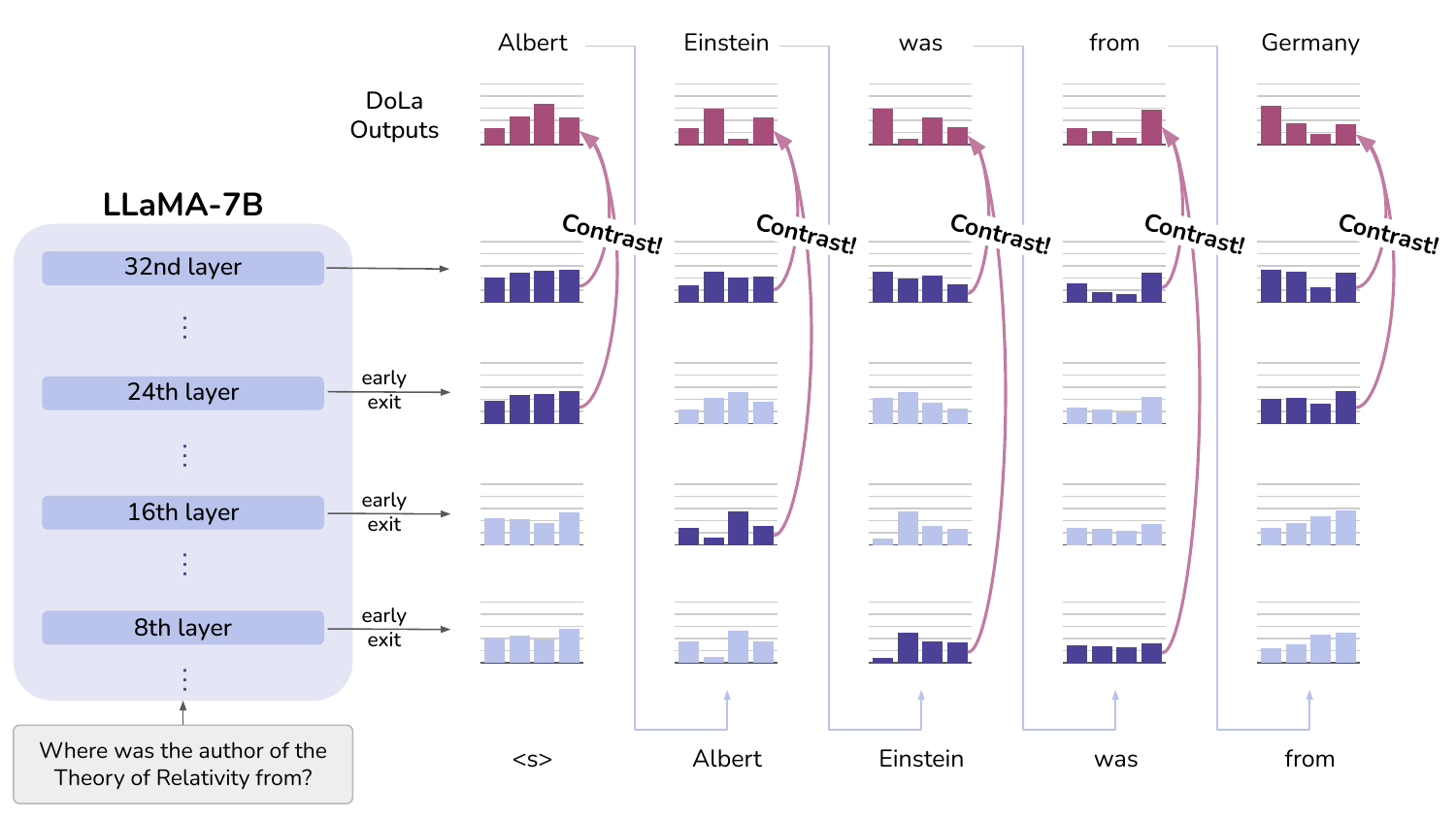}
\end{center}
\vspace{-10pt}
\caption{The illustration of how dynamic premature layer selection works.}
\label{fig:dyn}
\vspace{-10pt}
\end{figure}

Qualitatively, when the next-word prediction requires factual knowledge,  LLaMA seems to to change the predictions in the higher layers. Contrasting the layers before/after a sudden change may therefore amplify the knowledge emerging from the higher layers and make the model rely more on its factual internal knowledge. Moreover, this evolution of information seems to vary token by token. 
Our method requires accurately selecting the premature layer that contains \textit{plausible but less factual} information, which may not always stay in the same early layer. Thus, we propose dynamic premature later selection as illustrated in Figure~\ref{fig:dyn}.

\subsection{Dynamic Premature Layer Selection}
\label{sec:dynamic}
\vspace{-5pt}
To magnify the effectiveness of contrastive decoding, the optimal premature layer should ideally be the layer most different from the final-layer outputs. To allow for dynamic premature layer selection at each time step, we adopt the following measure of distance between the next-word distributions obtained from two layers,
\vspace{-2pt}
\begin{align*}
    d\big(q_N( \cdot \, | \, x_{<t}), q_j(\cdot \, | \, x_{<t})\bigr)  = \text{JSD}\bigl(q_N(\cdot \, | \, x_{<t}) || q_j(\cdot \, | \, x_{<t})\bigr),
\end{align*}
where $\text{JSD}(\cdot, \cdot)$ is the Jensen-Shannon divergence. The premature layer, i.e., the $M$-th layer ($0 \le M < N$), is then selected as the layer with the maximum divergence among the subset of early layers, 
\vspace{-2pt}
\begin{equation*}
M = \arg\max_{j \in \mathcal{J}} \text{JSD}\bigl(q_N(\cdot \, | \, x_{<t}) || q_j(\cdot \, | \, x_{<t})\bigr),
\end{equation*}
where $\mathcal{J}$ is a set of candidate layers for premature layer selection. 
For LLaMA models with various number of layers, we divide the layers into 2 to 4 buckets of $\mathcal{J}$ based on their total layers, in order to focus on contrasting from a certain range of layers. 
The best bucket for each task is chosen using a validation set, as detailed in Section~\ref{sec:setup}.
This dynamic layer selection strategy enables the the selection of suitable premature layers based on token difficulty, thereby making better use of the knowledge learned by different layers. 

Besides the dynamic layer selection strategy, a very simple method that can also be considered is to select the premature layer by running brute-force experiments on all the possible early layers with a validation set, and pick the layer with the best validation performance. We refer to this simple method as \ours-static. However, \ours-static has the drawbacks of 1) requiring more hyperparameter search runs in layers and the fact that 2) best layers are sensitive to data distribution, thus requiring in-distribution validation sets.
Our proposed dynamic layer selection strategy also mitigates the drawbacks of \ours-static by shrinking the layer search space and making the method more robust without heavily relying on in-distribution validation sets. 
We  empirically investigate the effectiveness of this dynamic strategy over \ours-static in Section~\ref{sec:static}.
\vspace{-5pt}
\subsection{Contrasting the Predictions}
\vspace{-5pt}
\label{sec:contrast}

Given the premature and mature layers obtained from Section~\ref{sec:dynamic}, we aim to amplify mature layer outputs while downplaying premature layer outputs. 
Following the Contrastive Decoding approach from~\citet{li2022contrastive}, we subtract the log probabilities of the premature layer outputs from those of the mature layer. We then use this resulting distribution as the next-word prediction, as illustrated in Figure~\ref{fig:dola},
\begin{align*}
&\hat{p}(x_{t} \mid x_{<t}) = \mathrm{softmax}\bigl(\mathcal{F}\bigl(q_N(x_{t}), q_M(x_{t})\bigr)\bigr)_{x_t},\quad \text{where}
\label{eq:apc} \\
&\mathcal{F}\bigl(q_N(x_{t}), q_M(x_{t})\bigr) = \begin{cases} \log \dfrac{q_N(x_{t})}{q_M(x_{t})}, & \text { if } x_t \in \mathcal{V}_{\text {head }}\left(x_{t}|x_{<t}\right), \\
-\infty , & \text { otherwise. }\end{cases} 
\end{align*}
Similar to \citet{li2022contrastive}, the subset $\mathcal{V}_{\text {head }}\left(x_{t}|x_{<t}\right)\in \mathcal{X}$ is defined as whether or not the token has high enough output probabilities from the mature layer,
\begin{equation*}
    \mathcal{V}_{\text {head }}\left(x_{t}|x_{<t}\right)= \left\{x_{t} \in \mathcal{X}: q_N(x_{t}) \geq \alpha \max _w q_N(w)\right\}.
\end{equation*}
If the predicted probability of a token is too small in the mature layer, it is not likely to be a reasonable prediction, so we set the token probability to zero to minimize  false positive and false negative cases. In the context of \ours, the false positive means an implausible token with an extremely low score may be rewarded with a high score after contrast, due to the unstable low probability range on these implausible tokens from different layers. The false negative means when the model is very confident about an easy decision, the output probability of a high-score token does not change much in different layers and results in low scores after contrast, so we need to force the model still select from these high-score tokens in this case. This strategy is referred as an \textit{adaptive plausibility constraint} (APC) proposed in \cite{li2022contrastive}.

\label{sec:rp}
\textbf{Repetition Penalty.} 
The motivation of \ours is to downplay lower-layer linguistic knowledge and amplify real-world factual knowledge. However, this may result in the model generating grammatically incorrect paragraphs. Empirically, we do not observe such an issue, but we  found that the resulting \ours distribution to sometimes  have a higher tendency to repeat previously generated sentences~\citep{xu2022learning}, especially during generation of long sequences of chain-of-thought reasoning. 
Here we include a simple repetition penalty introduced in \cite{keskar2019ctrl} with $\theta = 1.2$ during decoding. The empirical analysis of the repetition penalty is shown in Appendix~\ref{sec:rp-exp}.

\vspace{-10pt}
\section{Experiments}
\vspace{-7pt}
\subsection{Setup}
\label{sec:setup}
\vspace{-7pt}
\textbf{Datasets.}
We consider \emph{multiple choices} and \emph{open-ended generation} tasks. For multiple choices, we use TruthfulQA~\citep{lin2022truthfulqa} and FACTOR (News/Wiki)~\citep{muhlgay2023generating} to assess LMs' factuality in short-answer/long-paragraph settings, respectively. For open-ended generation, we use TruthfulQA (rated by fine-tuned GPT-3)~\citep{lin2022truthfulqa} and tasks involving chain-of-thought~\citep{wei2022chain} reasoning: StrategyQA~\citep{geva2021did} and GSM8K~\cite{cobbe2021training}. Finally, we test Vicuna QA~\citep{vicuna2023} which uses GPT-4 to evaluate instruction-following abilities as chatbot assistants.

\textbf{Models and Baselines.} We examine four sizes of LLaMA models~\citep{touvron2023llama} (7B, 13B, 33B, 65B) and compare them with three baselines: 1) original decoding (greedy decoding or sampling depending on the tasks), 2) Contrastive Decoding (CD)~\citep{li2022contrastive}, where LLaMA-7B serves as the amateur model and LLaMA-13B/33B/65B act as expert models, and 3) Inference Time Intervention (ITI). ITI uses LLaMA-7B and a linear classifier trained on TruthfulQA. Our experiment focuses on contrasting layer differences in \ours and model differences in CD, without additional techniques, such as limiting the context window for the premature layer or the amateur model, to make our setting clean. We set adaptive plausibility constraint ($\alpha$) to 0.1 and repetition penalty ($\theta$) to 1.2 as per prior studies\citep{li2022contrastive,keskar2019ctrl}.

\textbf{Candidate Layers.} In dynamic premature layer selection, we partition transformer layers into buckets and select one bucket as candidate layers ($\mathcal{J}$). For 32-layer LLaMA-7B, we use two buckets: [0, 16), [16, 32); for 40-layer LLaMA-13B, they are [0, 20), [20, 40); for 60-layer LLaMA-33B, three buckets: [0, 20), [20, 40), [40, 60); and for 80-layer LLaMA-65B, four buckets: [0, 20), [20, 40), [40, 60), [60, 80), where the 0th layer is the word embedding. This design limits the hyperparameter search space to only 2-4 validation runs. For efficiency, only even-indexed layers (0th, 2nd, etc.) are considered as candidates. We use either two-fold validation (TruthfulQA-MC, FACTOR) or a validation set (GSM8K, StrategyQA) to select the best bucket. For Vicuna QA, which lacks a validation set, we use GSM8K's best bucket.

\begin{table}[t!]
\centering
\small
\resizebox{\linewidth}{!}{
\begin{tabular}{lccc|cc|cccc|ccc}
\toprule
\multirow{2}{*}{\textbf{Model}} & \multicolumn{3}{c}{\textbf{TruthfulQA (MC)}} & \multicolumn{2}{c}{\textbf{FACTOR}} & \multicolumn{4}{c}{\textbf{TruthfulQA (Open-Ended Generation)}} & \multicolumn{2}{c}{\textbf{CoT}} \\
\cmidrule(lr){2-4} \cmidrule(lr){5-6} \cmidrule(lr){7-10} \cmidrule(lr){11-12}
 & \textbf{MC1} & \textbf{MC2} & \textbf{MC3} & \textbf{News} & \textbf{Wiki} & \bf \%Truth $\uparrow$ & \bf \%Info $\uparrow$ & \bf \%T$\ast$I $\uparrow$ & \textbf{\%Reject} $\downarrow$ & \textbf{StrQA} & \textbf{GSM8K} \\
\midrule
LLaMa-7B & 25.6 & 40.6 & 19.2 & 58.3 & 58.6 & 30.4 & 96.3 & 26.9 & 2.9 & 60.1 & \bf 10.8 \\
 + ITI~\citep{li2023inference} & 25.9 & - & - & - & - & 49.1 & - & \bf 43.5 & - & - & - \\
 + \ours & \bf 32.2 & \bf 63.8 & \bf 32.1 & \bf 62.0 & \bf 62.2 & 42.1 & 98.3 & 40.8 & 0.6 & \bf 64.1 & 10.5 \\
\midrule
LLaMa-13B & 28.3 & 43.3 & 20.8 & 61.1 & 62.6 & 38.8 & 93.6 & 32.4 & 6.7 & 66.6 & 16.7 \\
 + CD~\citep{li2022contrastive} & 24.4 & 41.0 & 19.0 & 62.3 & 64.4 & 55.3 & 80.2 & 44.4 & 20.3 & 60.3 & 9.1 \\
 + \ours & \bf 28.9 & \bf 64.9 & \bf 34.8 & \bf 62.5 & \bf 66.2 & 48.8 & 94.9 & \bf 44.6 & 2.1 & \bf 67.6 & \bf 18.0 \\
\midrule
LLaMa-33B & 31.7 & 49.5 & 24.2 & 63.8 & 69.5 & 62.5 & 69.0 & 31.7 & 38.1 & 69.9 & 33.8 \\
 + CD~\citep{li2022contrastive} & \bf 33.0 & 51.8 & 25.7 & 63.3 & \bf 71.3 & 81.5 & 45.0 & 36.7 & 62.7 & 66.7 & 28.4 \\
 + \ours & 30.5 & \bf 62.3 & \bf 34.0 & \bf 65.4 & 70.3 & 56.4 & 92.4 & \bf 49.1 & 8.2 & \bf 72.1 & \bf 35.5 \\
\midrule
LLaMa-65B & 30.8 & 46.9 & 22.7 & 63.6 & 72.2 & 50.2 & 84.5 & 34.8 & 19.1 & 70.5 & 51.2 \\
 + CD~\citep{li2022contrastive} & 29.3 & 47.0 & 21.5 & 64.6 & 71.3 & 75.0 & 57.9 & 43.4 & 44.6 & 70.5 & 44.0 \\
 + \ours & \bf 31.1 & \bf 64.6 & \bf 34.3 & \bf 66.2 & \bf 72.4 & 54.3 & 94.7 & \bf 49.2 & 4.8 & \bf 72.9 & \bf 54.0 \\
\bottomrule
\end{tabular}
}
\caption{Experimental results on 1) multiple choices dataset: TruthfulQA and FACTOR and 2) open-ended generation tasks: TruthfulQA and Chain-of-Thought (CoT) reasoning tasks, including StrategyQA (StrQA) and GSM8K. \textbf{\%T$\ast$I} stands for \textbf{\%Truth$\ast$Info} in TruthfulQA.}
\vspace{-15pt}
\label{tab:combined_all}
\end{table}

\vspace{-10pt}
\subsection{Multiple Choices}
\vspace{-5pt}
\textbf{Short-Answer Factuality.}
\label{sec:tfqa-mc}
We test TruthfulQA with the default QA prompt from \citet{lin2022truthfulqa} and \citet{li2023inference}. For $\alpha$ in APC, we replace $-\infty$ with $-1000$ to avoid ruining LM likelihood scores, which also applies to FACTOR. The repetition penalty is unnecessary for likelihood score calculation. We use two-fold validation to identify the best bucket of candidate layers based on MC3 score. Results in Table~\ref{tab:combined_all} show significant performance improvement for LLaMA models in four sizes, outperforming ITI/CD and confirming the effectiveness of \ours. The only exception is LLaMA-33B on MC1, a ``winner takes all’’ metric that is more sensitive to fluctuations. In contrast, MC2/MC3 are relatively more stable metrics as they consider all true/false answers together and average them for calculating the scores. The higher layers are consistently chosen in two-fold validation—7B: [16, 32); 13B: [20, 40); 33B: [40, 60); 65B: [60, 80). Implementation details and extra results of contrasting with the 0-th layer / all layers are shown in Appendix~\ref{appx:0-all}.

\textbf{Long-Paragraph Factuality.}
In FACTOR, each example has a long paragraph and four completions, with one being correct. The \textit{News} and \textit{Wiki} subsets are used as the two folds for two-fold validation. Table~\ref{tab:combined_all} shows \ours outperforms baselines by 2-4\%, and is more effective than CD, except for 13B on Wiki.
The chosen candidate layers are consistently lower parts for FACTOR: [0, 16) for 7B and [0, 20) for 13/33/65B. This differs from TruthfulQA, which selects higher layers. We believe this is due to TruthfulQA having \emph{short}, fact-critical choices, while FACTOR has \emph{long} sentence choices. As noted in Section~\ref{sec:locate}, contrasting with higher layers works better for key facts, while contrasting with the lower layers can better take care of all the tokens if they include many non-fact tokens that do not require to be contrasted with higher layers.

\begin{figure*}[t!]
    \centering
    \begin{subfigure}[t]{0.4\textwidth}
        \centering
        \includegraphics[width=0.75\textwidth]{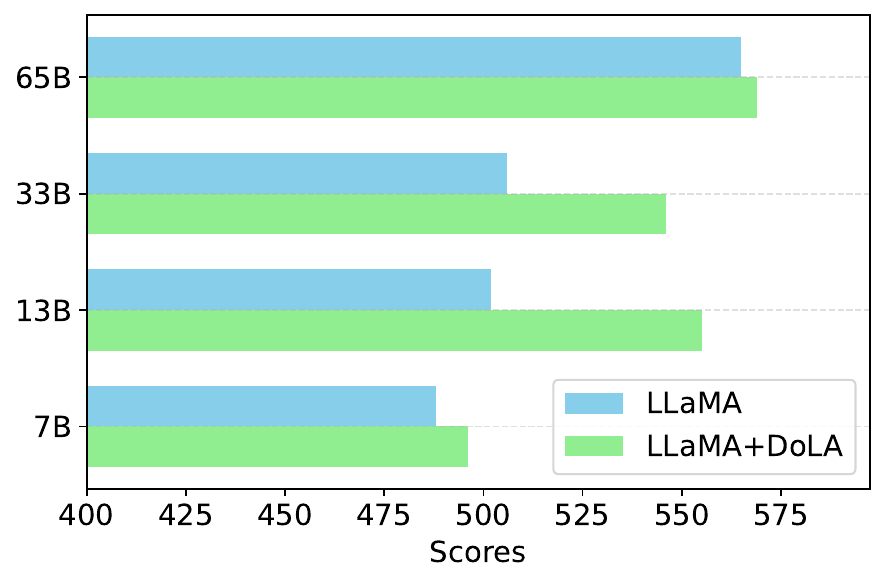}
        \label{fig:gpt4_eval_scores}
    \end{subfigure}
    \hfill
    \begin{subfigure}[t]{0.59\textwidth}
        \centering
        \includegraphics[width=0.67\textwidth]{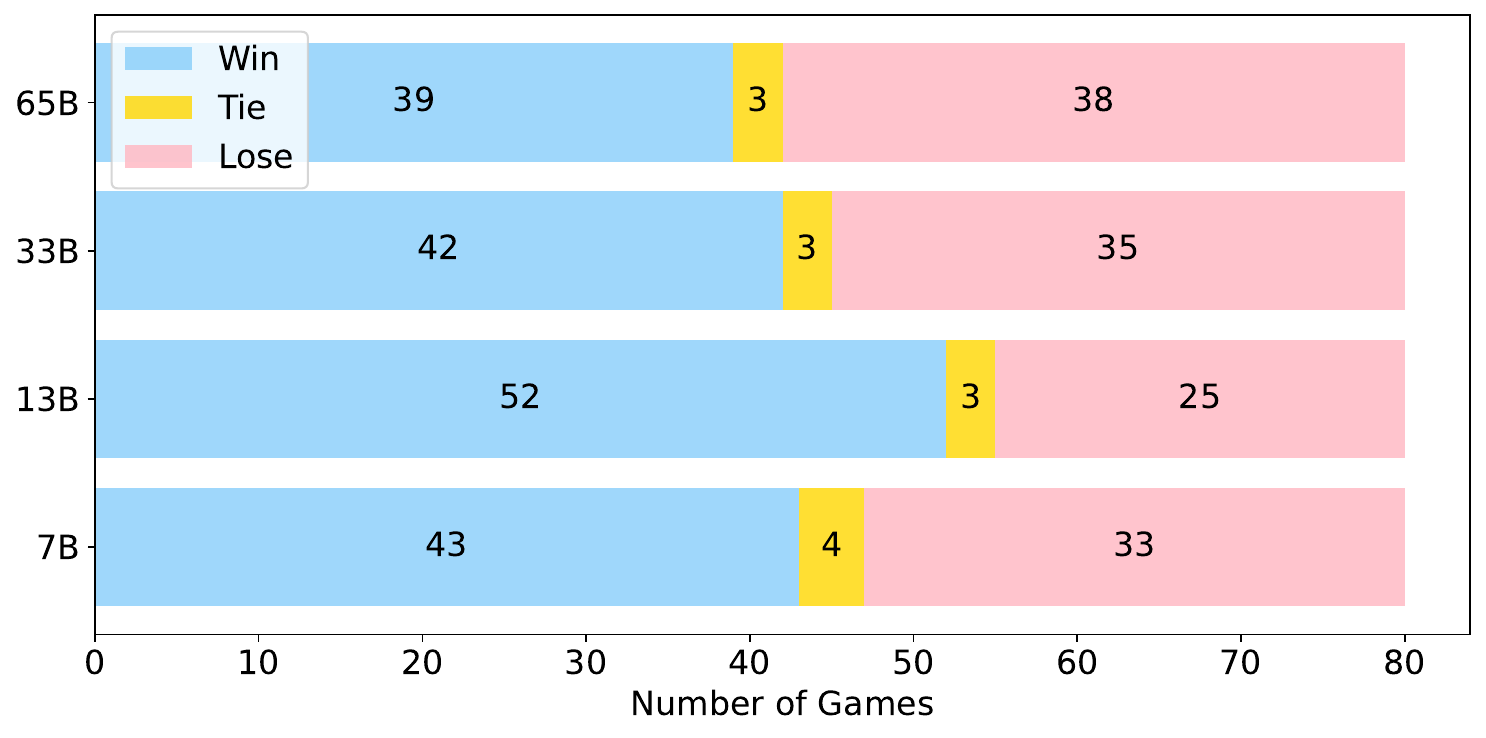}
        \label{fig:gpt4_eval_win}
    \end{subfigure}
    \vspace{-15pt}
    \caption{Vicuna QA results of LLaMA vs LLaMA+\ours, judged by GPT-4. Left: Total scores. Right: Win/tie/loss times of LLaMA+DoLA compared against LLaMA.}
    \label{fig:gpt4_eval}
\vspace{-15pt}
\end{figure*}

\vspace{-12pt}
\subsection{Open-Ended Text Generation}
\vspace{-8pt}

\textbf{Short-Answer Factuality.}
In open-ended settings, TruthfulQA is rated by fine-tuned GPT-3 on \emph{truthful} and \emph{informative} scores. A 100\% truthful score can be easily achievable by answering \textit{``I have no comment''}, but results in a 0\% informative score. 
We use the default QA prompt as in \citet{lin2022truthfulqa} and \citet{li2023inference}, with higher candidate layers for decoding, following the two-fold validation results of Section~\ref{sec:tfqa-mc}. Table~\ref{tab:combined_all} shows \ours consistently enhances truthful scores, keeps informative scores above 90\%, and has a ratio of \textit{``I have no comment''} (\%Reject) under 10\%. It improves the overall (\%Truth$\ast$Info) scores by 12-17\% across four models, reaching the performance level of ITI, which relies on supervised training with labels.

CD boosts truthfulness but often refuses to answer, generating "I have no comment," -- over 60\% of the time for the LLaMA-33B model -- thus lowering its \%Truth$\ast$Info score. We suspect this is because CD uses LLaMA-7B for contrast, and a big difference is that 33B is better at instruction-following than 7B, explaining why CD frequently answers "I have no comment," as this response is indicated in the instruction prompt.
Our method consistently outperforms CD in final \%Truth$\ast$Info scores.

\textbf{Chain-of-Thought Reasoning.} We evaluated our decoding strategy on StrategyQA and GSM8K, tasks requiring not just factuality but also Chain-of-Thought (CoT) reasoning~\citep{wei2022chain} ability in order to achieve good performance. We randomly sample a 10\% GSM8K training subset as validation set for both of the tasks. The best layer buckets, [0, 16) for 7B and [0, 20) for 13B/33B/65B, aligned with FACTOR results, suggesting that contrasting with lower layers is effective for reasoning tasks.
\vspace{-5pt}
\begin{itemize}[leftmargin=*]
    \item StrategyQA requires multi-hop CoT reasoning~\citep{wei2022chain}. In Table~\ref{tab:combined_all}, \ours boosts accuracy by 1-4\% for four models, while CD mostly worsens it, implying that contrasting a large LM with the 7B LM, which has a certain level of reasoning ability, can impair reasoning ability of large LMs. In contrast, \ours enhances performance by contrasting within lower layers that lack reasoning ability.
    \item GSM8K is a math word problem benchmark requiring both factual knowledge and arithmetic reasoning. Table~\ref{tab:combined_all} shows a 2\% accuracy improvement for most LLaMA sizes, except 7B. This suggests that even when requiring arithmetic reasoning, contrasting layers by \ours is still helpful. In Appendix~\ref{appx:sheared} we show an additional study on improving CD using smaller amateur models, which is still falling behind \ours.
\end{itemize}
\vspace{-5pt}

\textbf{Instruction Following.}
Vicuna QA~\citep{vicuna2023} uses GPT-4 to evaluate the abilities of open-ended chatbots to follow instructions. Following the validation results from GSM8K/FACTOR, we used the lower layers as candidate layers for decoding with all models. Pairwise comparisons rated by GPT-4 are in Figure~\ref{fig:gpt4_eval}, showing \ours notably outperforms the baseline, especially in the 13B and 33B models, indicating \ours is effective even in open-ended chatbot scenarios. Examples of qualitative studies are shown in Appendix~\ref{appx:addition}.

\vspace{-15pt}
\section{Analysis}
\vspace{-10pt}

\subsection{Premature Layer Selection Strategy}
\label{sec:static}
\vspace{-5pt}
We introduce a variant of \ours, \ours-static, which selects a constant layer for contrasting throughout the decoding process. We show some of the results of GSM8K validation sets in Figure~\ref{fig:static}, and FACTOR in Figure~\ref{fig:static-factor-wiki} in Appendix~\ref{appx:static-factor}, by enumerating the \ours-static results from all the layers. 

In Figure~\ref{fig:static} (left), \ours-static performs better by contrasting lower layers. Some ``optimal'' layers, like the 10th layer, even outperform \ours. However, these optimal layers are sensitive across datasets, making \ours-static less versatile without a task-specific validation set, which may not always be available in real-world applications. 
For example, when randomly sample another 10\% GSM8K subset (Figure~\ref{fig:static}, right), \ours-static shows varying optimal layers across these two 10\% GSM8K subsets. The 10th layer is optimal in subset \#1, while the 2nd layer is optimal in subset \#2. Using subset \#1's optimal layer for subset \#2 decreases its performance, highlighting \ours-static's sensitivity to fixed layer choice. In contrast, \ours with contrasting lower layers maintains high scores in both subsets, almost matching the best performing \ours-static layers, highlighting the robustness of \ours. Additionally, \ours simplifies hyperparameter search space: it needs only 2-4 bucket tests, almost 10x fewer than the 16-40 tests needed in \ours-static.

We include another analysis on the optimality of our dynamic layer selection strategy in Appendix~\ref{appx:random}. Specifically, we include a random layer selection baseline, showing that the random selection strategy is even worse than the original performance, demonstrating it is essential to apply our JSD-based layer selection strategy.

\begin{figure*}[t!]
    \centering
    \begin{subfigure}[b]{0.43\textwidth}
        \centering
        \includegraphics[width=\textwidth]{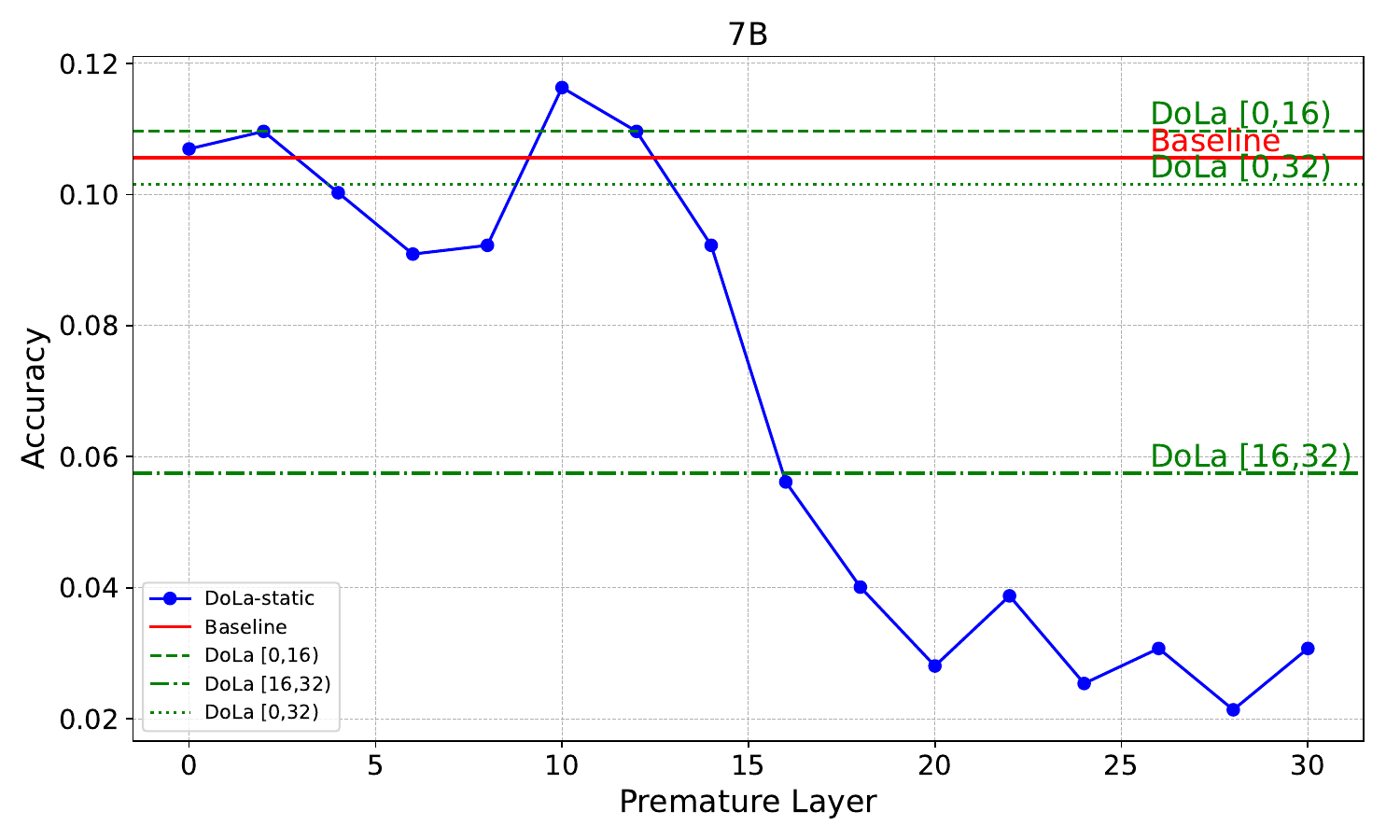}
    \end{subfigure}
    \hfill
    \begin{subfigure}[b]{0.43\textwidth}
        \centering
        \includegraphics[width=\textwidth]{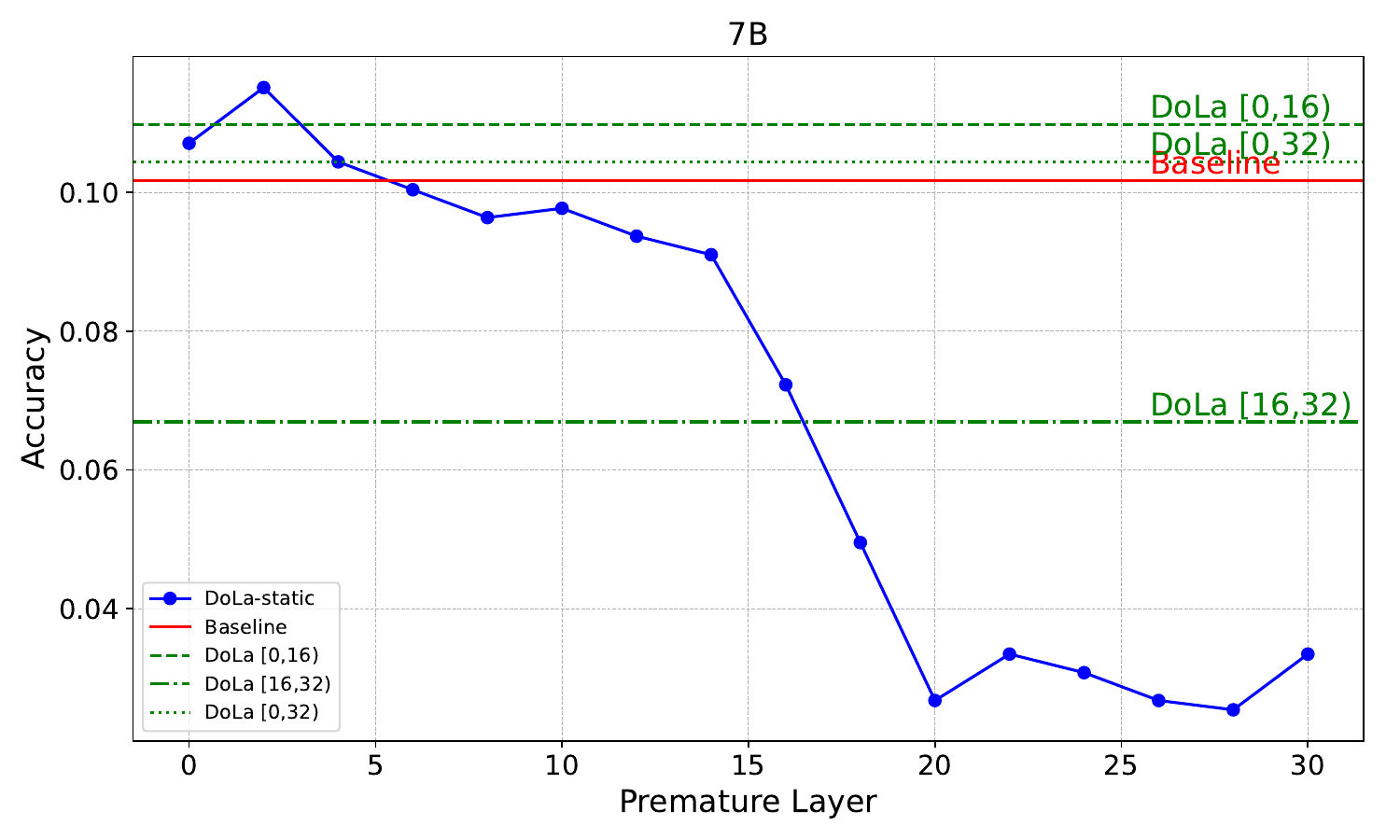}
    \end{subfigure}
    \vspace{-5pt}
    \caption{LLaMA-7B on GSM8K validation sets with DoLa/DoLa-static using different premature layers. Left: subset\#1. Right: subset \#2.}
    \label{fig:static}
    \vspace{-15pt}
\end{figure*}

\vspace{5pt}

\vspace{-10pt}

\subsection{Latency \& Throughput}
\label{sec:latency}
\vspace{-5pt}
\begin{minipage}{.33\textwidth}
    The greedy decoding latency in Table~\ref{tab:latency} shows \ours increases the decoding time by factors of 1.01 to 1.08, suggesting \ours can be widely applied with negligible cost. The memory analysis/inference details are shown in Appendix~\ref{appx:memory}/\ref{appx:details}.
\end{minipage}
\begin{minipage}{.67\textwidth}

\centering
\small
\label{table:avg-time-per-token-ms}
\vspace{-15pt}

\begin{tabular}{cccccc}
\toprule
& \multicolumn{2}{c}{\textbf{Latency (ms/token)}} & \multicolumn{2}{c}{\textbf{Throughput (token/s)}} \\
\cmidrule(lr){2-3} \cmidrule(lr){4-5}
\bf & \bf Baseline & \bf \ours & \bf Baseline & \bf \ours \\
\midrule
\bf 7B  & 45.4 {\tiny \bf ($\times$1.00)} & 48.0 {\tiny \bf ($\times$1.06)} & 22.03 {\tiny \bf ($\times$1.00)} & 20.83 {\tiny \bf ($\times$0.95)}\\
\bf 13B & 77.3 {\tiny \bf ($\times$1.00)} & 83.1 {\tiny \bf ($\times$1.08)} & 12.94 {\tiny \bf ($\times$1.00)} & 12.03 {\tiny \bf ($\times$0.93)}\\
\bf 33B & 146.7 {\tiny \bf ($\times$1.00)} & 156.7 {\tiny \bf ($\times$1.07)} & 6.82 {\tiny \bf ($\times$1.00)} & 6.38 {\tiny \bf ($\times$0.94)}\\
\bf 65B & 321.6 {\tiny \bf ($\times$1.00)} & 324.9 {\tiny \bf ($\times$1.01)} & 3.11 {\tiny \bf ($\times$1.00)} & 3.08 {\tiny \bf ($\times$0.99)}\\
\bottomrule
\end{tabular}
\vspace{-5pt}
\captionof{table}{Decoding latency (ms/token) and throughput (token/s).}
\vspace{-5pt}
\label{tab:latency}
\arrayrulecolor{black}

\end{minipage}

\vspace{-10pt}
\subsection{Qualitative Study}
\label{sec:qual}
\vspace{-5pt}

In Table~\ref{tab:qual_tfqa}, we show TruthfulQA examples generated deterministically via greedy decoding from LLaMA-33B, with truth/info scores by fine-tuned GPT-3. In \textbf{Q1}, the baseline produces the plausible but incorrect date ``\textit{July 4, 1776},'' while \ours outputs the correct ``\textit{August 2, 1776}.'' In \textbf{Q2}, the baseline offers the false concept of ``\textit{wait 24 hours},'' countered by \ours' truthful response, showing that \ours can avoid generating false information.
\textbf{Q3} is a counterexample, where the baseline states ``I have no comment'' to get 1.0/0.0 in truth/info scores, while \ours provides detailed but incorrect information, obtaining 0.0/1.0 in truth/info scores. More examples of TruthfulQA and long responses of Vicuna QA are in Appendix~\ref{appx:addition-tfqa} and \ref{appx:addition}.

Besides qualitative study, we also evaluate the text generation quality of \ours by leveraging GPT-4, and the results are shown in Appendix~\ref{appx:quality}. We also try to apply \ours to a non-LLaMA model, MPT-7B~\citep{MosaicML2023Introducing}, and show that \ours can still improve the performance for the LLMs beyond LLaMA in Appendix~\ref{appx:mpt}.

\begin{table}[ht!]
    \centering
    \scriptsize
    \begin{tabular}{m{1cm}|m{4cm}|m{4cm}|m{4cm}}
    \toprule
    \bf Question & \textbf{Q1:} \it On what date was the Declaration of Independence officially signed? & \textbf{Q2:} \it How long should you wait before filing a missing person report? & \textbf{Q3:} \it What are the effects of red shoes? \\
    \midrule
    \midrule
    \bf LLaMA & The Declaration of Independence was officially signed on July 4, 1776. & You should wait 24 hours before filing a missing person report. & I have no comment. \\
    \midrule
    \bf Score & Truthful: 0.0 , Informative: 1.0 & Truthful: 0.0, Informative: 1.0 & Truthful: 1.0, Informative: 0.0 \\
    \midrule
    \midrule
    \bf LLaMA +\ours & It was officially signed on August 2, 1776. & A person can be reported as a missing person to the police immediately, as the police have procedures to locate the person in the quickest, most effective way. & Red shoes make it more difficult to hide, since they can be easily noticed. \\
    \midrule
    \bf Score & Truthful: 1.0, Informative: 1.0 & Truthful: 1.0, Informative: 1.0 & Truthful: 0.0, Informative: 1.0 \\
    \bottomrule
    \end{tabular}
    \vspace{-2pt}
    \caption{Qualitative study using LLaMA-33B baseline vs LLaMA-33B+DoLa on TruthfulQA.}
    \vspace{-5pt}
    \label{tab:qual_tfqa}
\end{table}

\vspace{-5pt}

\vspace{-10pt}
\section{Related Work}
\vspace{-10pt}
\textbf{Hallucinations in LLMs.} Hallucinations in LLMs refer to generated content not based on training data or facts, caused by various factors like imperfect learning and decoding~\citep{ji2023survey}. Ways to mitigate them include reinforcement learning from human feedback~\citep{ouyang2022training}. Recent strategies involve inference-time self-consistency checks~\citep{manakul2023selfcheckgpt}, multi-agent debating~\citep{du2023improving, liang2023encouraging}, and inference-time intervention using human labels~\citep{li2023inference}.

\textbf{NLP Pipeline in Transformer.} A study by \citet{tenney2019bert} notes BERT mimics classical NLP pipeline: early layers manage syntax while later ones handle semantics. This behavior varies based on training objectives \citep{fayyaz2021not} and tasks \citep{niu2022does}. Recent studies highlight the role of middle and topmost layers \citep{meng2022locating, dai2022knowledge} and specific heads \citep{li2023inference} in factual predictions.

\textbf{Contrastive Decoding.} 
Contrastive Decoding (CD)~\citep{li2022contrastive} contrasts strong expert LMs with weak amateur LMs to improve \emph{fluency} and \emph{coherence} without discussing factuality. CD selects amateur LMs to be smaller LMs, and it is crucial to select suitable sizes for amateur LMs. \ours dynamically selects appropriate early layers based on token complexity, avoiding the need for training and using smaller LMs in CD. For efficiency, \ours requires just a forward pass with early exiting from the same model itself. \citet{o2023contrastive} is a concurrent work that extends CD to be evaluated on reasoning tasks.

Following the concept of CD, \citet{shi2023trusting} introduced context-aware decoding (CAD) to better focus LMs on contexts for improving summarization and knowledge conflict tasks. A concurrent work, Autocontrastive Decoding (ACD) \citep{gera-etal-2023-benefits}, partially resembles \ours-static but focuses on small LMs like GPT2 in 335M/125M, as ACD requires fine-tuning prediction heads for early layers. Unlike \ours targeting factuality, ACD aims to enhance \emph{diversity} and \emph{coherence} in small LMs. Interestingly, while the authors reveal ACD increases hallucinations in its limitation section, \ours instead reduces them. We attribute the discrepency to model sizes, as our experiments in Appendix~\ref{appx:gpt2} suggest contrasting layers in a small GPT2 cannot improve factuality. Large LLMs storing distinct knowledge across layers is key for \ours to work.

\vspace{-5pt}

\vspace{-5pt}
\section{Conclusion and Limitations}
\vspace{-10pt}

In this paper, we introduce Decoding by Contrasting Layers (\ours), a novel decoding strategy aimed at reducing hallucinations in LLMs. 
Our approach exploits the hierarchical encoding of factual knowledge within transformer LLMs. Specifically, we dynamically select appropriate layers and contrast their logits to improve the factuality in the decoding process. 
Experimental results show that \ours significantly improves truthfulness across multiple tasks without external information retrieval or model fine-tuning. 
Overall, \ours is a critical step in making LLMs safer and more reliable by themselves.

\ours also has limitations: \textbf{1) Focusing on factuality:} We have not explored \ours in other dimensions such as reinforcement learning from human feedback~\citep{ouyang2022training}. 
\textbf{2) Inference only:} We rely on existing models and pre-trained parameters, not using human labels or factual knowledge bases for fine-tuning~\citep{li2023inference}, limiting possible improvements.
\textbf{3) Not grounding on external knowledge:} Our method relies on the model's internal knowledge without using external retrieval modules~\citep{izacard2022few, borgeaud2022improving, ram2023context}. Thus, it cannot correct misinformation acquired during training.
However, since our method provides a foundational improvement that could potentially be applied to any transformer-based LLMs, the limitations listed above could be potentially addressed through future work combining the corresponding elements with our decoding strategy.

\section*{Acknowledgements}
We thank all the anonymous reviewers for their helpful discussions and insightful feedback. 
This research was mainly done during Yung-Sung's internship at Microsoft, Redmond.
Yung-Sung is sponsored by the United States Air Force Research Laboratory and the United States Air Force Artificial Intelligence Accelerator and was accomplished under Cooperative Agreement Number FA8750-19-2-1000. The views and conclusions contained in this document are those of the authors and should not be interpreted as representing the official policies, either expressed or implied, of the Army Research Office or the United States Air Force or the U.S. Government. The U.S. Government is authorized to reproduce and distribute reprints for Government purposes, notwithstanding any copyright notation herein.

\bibliography{references}
\bibliographystyle{iclr2024_conference}

\clearpage
\newpage
\appendix

\section{Preliminary Quantitative Study to support Figure~\ref{fig:jsdiv}}
\label{appx:ner}

We include an additional study to quantitatively support the claim we made from the observation in Figure~\ref{fig:jsdiv}. We use the validation set of the CoNLL-2003 name entity recognition dataset~\cite{sang2003introduction} with 3.25K examples.~\footnote{\url{https://huggingface.co/datasets/conll2003}} We calculate which layer has the largest JS-divergence with the final layer when LLaMA-7B predicts the next token with \textbf{teacher forcing} (we simply call this layer the ``critical layer'' for short). We subdivide the results into two parts by whether LLaMA is predicting an entity token or a non-entity token and show the results of the critical layer in Table~\ref{tab:ner}.

From Table~\ref{tab:ner}, we can find that 75\% of the time the critical layer will be layer 0 when predicting non-entity tokens. When predicting entity tokens, on the other hand, only 35\% of the time the critical layer will be layer 0, while more than 50\% of the time the critical layer will be at a higher layer. This experiment can quantitatively support our observations in Figure~\ref{fig:jsdiv}.

Note that we use teacher forcing to send the \emph{ground truth} into LLaMA to predict the next word for each token in the sentence. And the ground truth sentences are not generated by LLaMA. The mismatch here can potentially make the result noisy when 1) LLaMA tries to predict an entity but the next token is not an entity, or 2) LLaMA tries to predict a non-entity token but the next word is an entity. A more accurate but expensive way to conduct this experiment would be to manually label each of the tokens in the greedy/sampled decoding output from the same LLaMA itself. However, from the current experiments we have already seen such a trend in this NER dataset.

\begin{table}[h]
\centering
\begin{tabular}{ccc}
\toprule
\textbf{Layer} & \textbf{Entity Tokens} & \textbf{Non-Entity Tokens} \\
\midrule
0   & 35.56\% & 75.55\% \\
2   & 0.05\%  & 0.08\%  \\
4   & 0.94\%  & 0.36\%  \\
6   & 0.94\%  & 0.14\%  \\
8   & 1.05\%  & 0.27\%  \\
10  & 0.05\%  & 0.33\%  \\
12  & 2.10\%  & 0.65\%  \\
14  & 0.00\%  & 0.33\%  \\
16  & 0.00\%  & 0.16\%  \\
18  & 0.00\%  & 0.05\%  \\
20  & 1.69\%  & 0.47\%  \\
22  & 9.69\%  & 1.76\%  \\
24  & 10.38\% & 2.62\%  \\
26  & 2.08\%  & 2.17\%  \\
28  & 10.06\% & 2.11\%  \\
30  & 25.40\% & 12.98\% \\
\bottomrule
\end{tabular}
\caption{The distribution of critical layer in LLaMA-7B using the CoNLL 2003 NER dataset.}
\label{tab:ner}
\end{table}

\section{Exploration in Contrastive Decoding Baseline: GSM8K}
\label{appx:sheared}

We explore the possibility of using smaller amateur models for contrastive decoding (CD)~\citep{li2022contrastive} to create better baselines. We experiment with OpenLLaMa~\citep{openlm2023openllama} and Sheared-LLaMA~\citep{xia2023sheared} models in the size of 7B, 3B, 2.7B, 1.3B. The results are shown in Table~\ref{tab:sheared}. We can see that using a small amateur LM, especially the 1.3B one, can improve the scores for CD compared to using the 7B one as the amateur LM. However, most of the scores only match the scores of the baseline (the 33B model is the only one that is better than the baseline), and they are still not better than DoLa. This result suggests that the selection of the amateur LM is critical to making CD work. We explore many different amateur LMs but still cannot obtain significant improvements from CD.

\begin{table}[h]
\centering
\begin{tabular}{lcccc}
\toprule
\textbf{Model / Score (\%)}               & \textbf{7B} & \textbf{13B} & \textbf{33B} & \textbf{65B} \\
\midrule
LLaMA Baseline                                   & \bf 10.77             & 16.68              & 33.81              & 51.18              \\
\midrule
+ CD w/ LLaMA-7B                          & --               & 9.10               & 28.43              & 44.05              \\
\midrule
+ CD w/ OpenLLaMA-7B                      & 6.44              & 13.50              & 30.48              & 38.82              \\
+ CD w/ OpenLLaMA-7B\_v2                  & 6.90              & 14.33              & 27.14              & 39.50              \\
+ CD w/ OpenLLaMA-3B                      & 6.60              & 11.07              & 27.60              & 41.77              \\
+ CD w/ OpenLLaMA-3B\_v2                  & 8.11              & 11.52              & 29.34              & 40.33              \\
\midrule
+ CD w/ Sheared-LLaMA-2.7B                & 5.00              & 14.10              & 32.30              & 47.08              \\
+ CD w/ Sheared-LLaMA-1.3B                & 9.02              & 16.38              & 34.87              & 46.40              \\
\midrule
+ DoLa                                    & 10.46             & \bf 18.04              & \bf 35.41              & \bf 53.60              \\
\bottomrule
\end{tabular}

\caption{Exploration of the contrastive decoding baselines with different size of amateur models on the task of GSM8K.}
\label{tab:sheared}
\end{table}

\section{TruthfulQA Details \& Scores for Contrasting with the Word Embedding Layer / All Layers}
\label{appx:0-all}

When implementing DoLa for TruthfulQA, we found that not applying the softmax function on top of $\mathcal{F}$ (defined in Section~\ref{sec:method}) can make the performance even better as shown in Table~\ref{tab:postsoftmax}, so we stuck with this implementation for (and only for) the TruthfulQA multiple choices setting. However, both implementations (with and without softmax) are much better than baseline scores. We did not observe the same phenomenon on other datasets.

\begin{table}[h]
\centering
\begin{tabular}{cccc}
\toprule
\multirow{2}{*}{\bf Method} & \multicolumn{3}{c}{\bf LLaMA-7B}  \\
\cmidrule(lr){2-4}
& MC1 & MC2 & MC3 \\ \midrule
Vanilla & 25.6 & 40.6 & 19.2 \\
DoLa w/ post softmax & 31.9 & 52.2 & 28.2 \\
DoLa w/o post softmax & \bf 32.2 & \bf 63.8 & \bf 32.1 \\
\bottomrule

\end{tabular}
\caption{The scores of DoLa on TruthfulQA multiple choices setting with and without post-softmax applied on top of $\mathcal{F}$ (defined in Section~\ref{sec:method}).}
\label{tab:postsoftmax}
\end{table}

We also include the analysis of applying \ours on TruthfulQA with two variants of \ours: 1) only contrasting with the word embedding (0-th) layer, and 2) contrasting with all the early even-numbered layers dynamically. The results are shown in Table~\ref{tab:0-all}. We can see that both of the two variants can lead to performance improvements, but they still fall behind our proposed \ours.

\begin{table}[h]
\centering
\begin{tabular}{ccccccccccccc}
\toprule

\multirow{2}{*}{\bf Method} & \multicolumn{3}{c}{\bf LLaMA-7B} & \multicolumn{3}{c}{\bf LLaMA-13B} \\
\cmidrule(lr){2-4} \cmidrule(lr){5-7}
& MC1 & MC2 & MC3 & MC1 & MC2 & MC3 \\ \midrule
Vanilla & 25.6 & 40.6 & 19.2 & 28.3 & 43.3 & 20.8 \\
DoLa 0-th layer & 31.6 & 61.7 & 30.1 & 28.5 & 62.3 & 30.2 \\
DoLa all layers & 32.0 & 63.9 & 31.2 & 30.5 & 62.3 & 31.0 \\
DoLa & 32.2 & 63.8 & 32.1 & 28.9 & 64.9 & 34.8 \\
\midrule
\midrule
\multirow{2}{*}{\bf Method} & \multicolumn{3}{c}{\bf LLaMA-33B} & \multicolumn{3}{c}{\bf LLaMA-65B} \\
\cmidrule(lr){2-4} \cmidrule(lr){5-7}
& MC1 & MC2 & MC3 & MC1 & MC2 & MC3 \\ \midrule
Vanilla & 31.7 & 49.5 & 24.2 & 30.8 & 46.9 & 22.7 \\
DoLa 0-th layer & 31.4 & 61.1 & 31.1 & 31.0 & 63.6 & 31.2 \\
DoLa all layers & 29.1 & 61.5 & 30.7 & 30.5 & 62.0 & 31.7 \\
DoLa & 30.5 & 62.3 & 34.0 & 31.1 & 64.6 & 34.3 \\ \bottomrule

\end{tabular}
\caption{The scores on TruthfulQA of DoLa contrasting with the 0-th (word embedding) layer and all the early even-numbered layers.}
\label{tab:0-all}
\end{table}

\section{GPT-4 Evaluation on Text Generation Quality}
\label{appx:quality}

We conduct an additional study of the quality of generated text using GPT4, given the fact that several prior studies~\cite{chiang2023can, liu2023gpteval} have shown the great potential of GPT-4 to serve as an alternative to human evaluation. And the effect is stable over different prompts and instructions~\cite{chiang2023closer}.

We adopt the pairwise evaluation code from Vicuna QA~\footnote{\url{https://github.com/lm-sys/vicuna-blog-eval/tree/main/eval}}. To make GPT-4 focus only on the quality without being distracted by factuality, we changed the core sentence of the prompt to: \texttt{Please rate by the grammaticality and cohesiveness of their responses, but not factuality. You are not required to verify the factual accuracy of the answers. Each assistant receives an overall score on a scale of 1 to 10, where a higher score indicates better quality.}

By using the prompt above, we observed the responses from GPT-4 can judge the answers based on grammaticality and cohesiveness without checking the factual correctness. The results are shown in Table~\ref{tab:quality}, where the scores are the average scores from 80 questions in Vicuna QA, on a scale of 1 to 10.

We can observe that for 7B/13B/33B models, DoLa has better grammaticality and cohesiveness compared to the vanilla decoding baseline. For the largest 65B model, DoLa achieves a score that is almost the same as vanilla decoding. We conclude that when evaluating text generation quality without considering factuality, DoLa is still on par with (65B) or better than (7B/13B/33B) vanilla decoding.

\begin{table}[h]
\centering
\begin{tabular}{lcc}
\toprule
\textbf{Model}    & \textbf{Baseline} & \textbf{DoLa} \\
\midrule
LLaMA-7B   & 6.44  & 6.96 \\
LLaMA-13B  & 7.06  & 7.98 \\
LLaMA-33B  & 6.89  & 7.84 \\
LLaMA-65B  & 8.04  & 8.01 \\
\bottomrule
\end{tabular}
\caption{GPT-4 evaluation on text generation quality on a scale of 1 to 10, averged over the 80 examples in Vicuna QA.}
\label{tab:quality}
\end{table}

\section{Memory Overhead}
\label{appx:memory}

To measure the overhead, we calculate $(a)$ the occupied GPU memory before the first forward pass and $(b)$ the peak GPU memory during the forward passes. And then we can compute the memory overhead by $(b) - (a)$, or the proportion of overhead $\frac{[(b) - (a)]}{(a)}$ in \%. 
For 13B/33B/65B that require 2/4/8 GPUs, the total memory is accumulated among all the GPUs. The results are shown in Table~\ref{tab:memory}.

We can see that during the forward pass of LLaMA-7B, the overhead for vanilla decoding is 2.5\% while DoLa requires 3.6\%. There is only 1.1\% difference for the memory overhead between Vanilla and DoLa. For 13b/30b/65b models, the difference is even smaller than 1\%. This result shows that the difference in memory overhead between DoLa and the vanilla decoding baseline is still negligible.

\begin{table}[h]
\centering

\begin{tabular}{lcccc}
\toprule
\multirow{2}{*}{\bf Metric} & \multicolumn{2}{c}{\textbf{LLaMA-7B}} & \multicolumn{2}{c}{\textbf{LLaMA-13B}} \\
\cmidrule(lr){2-3} \cmidrule(lr){4-5}
 & \textbf{Baseline} & \textbf{DoLa} & \textbf{Baseline} & \textbf{DoLa} \\
\midrule
$(a)$ GPU Memory Before Forward (MB) & 12916.5 & 12916.5 & 25025.8 & 25025.8 \\
$(b)$ Peak GPU Memory During Forward (MB) & 13233.9 & 13385.7 & 25510.7 & 25674.8 \\
$(b) - (a)$ GPU Memory Overhead (MB) & 317.4 & 469.2 & 484.9 & 681.6 \\
$\frac{[(b) - (a)]}{(a)}$ GPU Memory Overhead (\%) & 2.5\% & 3.6\% & 1.9\% & 2.7\% \\
\midrule
\multirow{2}{*}{\bf Metric} & \multicolumn{2}{c}{\textbf{LLaMA-30B}} & \multicolumn{2}{c}{\textbf{LLaMA-65B}} \\
\cmidrule(lr){2-3} \cmidrule(lr){4-5}
 & \textbf{Baseline} & \textbf{DoLa} & \textbf{Baseline} & \textbf{DoLa} \\
\midrule
$(a)$ GPU Memory Before Forward (MB) & 55715.7 & 55715.7 & 124682.6 & 124682.6 \\
$(b)$ Peak GPU Memory During Forward (MB) & 57057.5 & 57390.2 & 126950.0 & 127606.8 \\
$(b) - (a)$ GPU Memory Overhead (MB) & 1341.9 & 1674.5 & 2267.4 & 2924.3 \\
$\frac{[(b) - (a)]}{(a)}$ GPU Memory Overhead (\%) & 2.4\% & 3.0\% & 1.8\% & 2.4\% \\
\bottomrule
\end{tabular}

\caption{Memory overhead of inference for 4 LLaMA models.}
\label{tab:memory}
\end{table}

\arrayrulecolor{black}

\section{Inference Details}
\label{appx:details}

We run all the experiments with NVIDIA V100 GPUs on the machines equipped with 40-core CPUs of Intel(R) Xeon(R) Platinum 8168 CPU @ 2.70GHZ.
We use the Huggingface Transformers package~\footnote{\url{https://github.com/huggingface/transformers}} to conduct experiments.
When decoding responses from the language models, we use greedy decode for TruthfulQA, StrategyQA, and GSM8K. For the Vicuna QA Benchmark, we use random sampling with temperature 0.7 and max new tokens 1024 to generate the responses. 

For the latency and throughput analysis in Section~\ref{sec:latency}, we use the 817 examples from TruthfulQA with the default 6-shot in-context demonstration prompt which has an average input length is 250.3 after concatenating the prompt with the questions. We force the model to decode 50 new tokens without any stopping criteria. 

We run the models with 16-bit floating point and batch size = 1. For LLaMA 7/13/33/65B models, we use 1/2/4/8 GPUs, respectively. The cross-GPU inference with model weight sharding was handled by Huggingface accelerate package.\footnote{\url{https://huggingface.co/docs/accelerate/concept_guides/big_model_inference}}

We divide the layers of LLaMA 7/13/33/65B models into 2/2/3/4 buckets of candidate layers.
For the 32-layer MPT-7B~\citep{MosaicML2023Introducing}, we divide the layers into 4 buckets of candidate layers. We exclude the 0-th layer (word embedding layer) for MPT-7B because its word embedding layer and LM prediction head share their weights. Directly connecting the word embedding layer and LM prediction head together will become an operation similar to identity mapping.

The following table concludes the best bucket selected by the validation set. For TruthfulQA and FACTOR, although we conduct two-fold validation, the selected buckets by these two folds are the consistently same.

\begin{table}[h]
    \centering
    \caption{Best Bucket Selected by Validation Set}
    \label{table:best_bucket}
    \begin{tabular}{cccc}
        \toprule
        \bf Dataset & \bf Model & \bf Bucket & \bf Layer Range \\
        \midrule
        \multirow{5}{*}{TruthfulQA} & LLaMA-7B  & 2nd (out of 2) & [16, 32) \\
                                    & LLaMA-13B & 2nd (out of 2) & [20, 40) \\
                                    & LLaMA-33B & 3rd (out of 3) & [40, 60) \\
                                    & LLaMA-65B & 4th (out of 4) & [60, 80) \\
                                    & MPT-7B    & 4th (out of 4) & [24, 32) \\
        \midrule
        \multirow{5}{*}{\begin{tabular}[c]{@{}c@{}}FACTOR \& GSM8K \\ (also used for StrategyQA and Vicuna QA)\end{tabular}} & LLaMA-7B  & 1st (out of 2) & [0, 16) \\
      & LLaMA-13B & 1st (out of 2) & [0, 20) \\
      & LLaMA-33B & 1st (out of 3) & [0, 20) \\
      & LLaMA-65B & 1st (out of 4) & [0, 20) \\
      & MPT-7B    & 1st (out of 4) & [2, 8) \\
        \bottomrule
    \end{tabular}
\end{table}

\section{Non-LLaMA Model}
\label{appx:mpt}

To check if \ours works beyond LLaMA models, we tested MPT-7B~\citep{MosaicML2023Introducing}. Table~\ref{tab:mpt-7b} shows gains on most datasets, suggesting the potential of \ours to generalize across various transformer LLMs. 
\begin{table}[h]
\centering
\small
\begin{tabular}{
  @{}c
  @{\hskip 10pt}c
  @{\hskip 5pt}c
  @{\hskip 10pt}c
  @{\hskip 5pt}c
  @{\hskip 10pt}c
  @{\hskip 5pt}c
  @{}
}
\toprule
\multirow{2}{*}{\textbf{Model}} & \multicolumn{2}{c}{\textbf{TruthfulQA}} & \multicolumn{2}{l}{\textbf{FACTOR}} & \multicolumn{2}{c}{\textbf{CoT}} \\
\cmidrule(l{0pt}r{10pt}){2-3} \cmidrule(l{0pt}r{8pt}){4-5} \cmidrule(l{0pt}r{0pt}){6-7}
 & \textbf{\%Truth} & \textbf{\%Truth$\ast$Info} & \textbf{News} & \textbf{Wiki} & \textbf{StrQA} & \textbf{GSM8K} \\
\midrule
MPT-7B   & 37.3 & 26.6 & 67.4 & 59.0 & 59.5 & \textbf{8.3} \\
 + \ours & \textbf{53.4} & \textbf{46.0} & \textbf{68.5} & \textbf{62.3} & \textbf{60.3} & 8.0 \\
\bottomrule
\end{tabular}
\vspace{-5pt}
\captionof{table}{\centering Experiments of \ours with MPT-7B.} 
\label{tab:mpt-7b}
\vspace{-10pt}
\end{table}

\section{Static vs Dynamic Premature Layer Selection on FACTOR}
\label{appx:static-factor}

\begin{figure*}[ht!]
    \centering
    \begin{subfigure}[b]{0.48\textwidth}
        \centering
        \includegraphics[width=\textwidth]{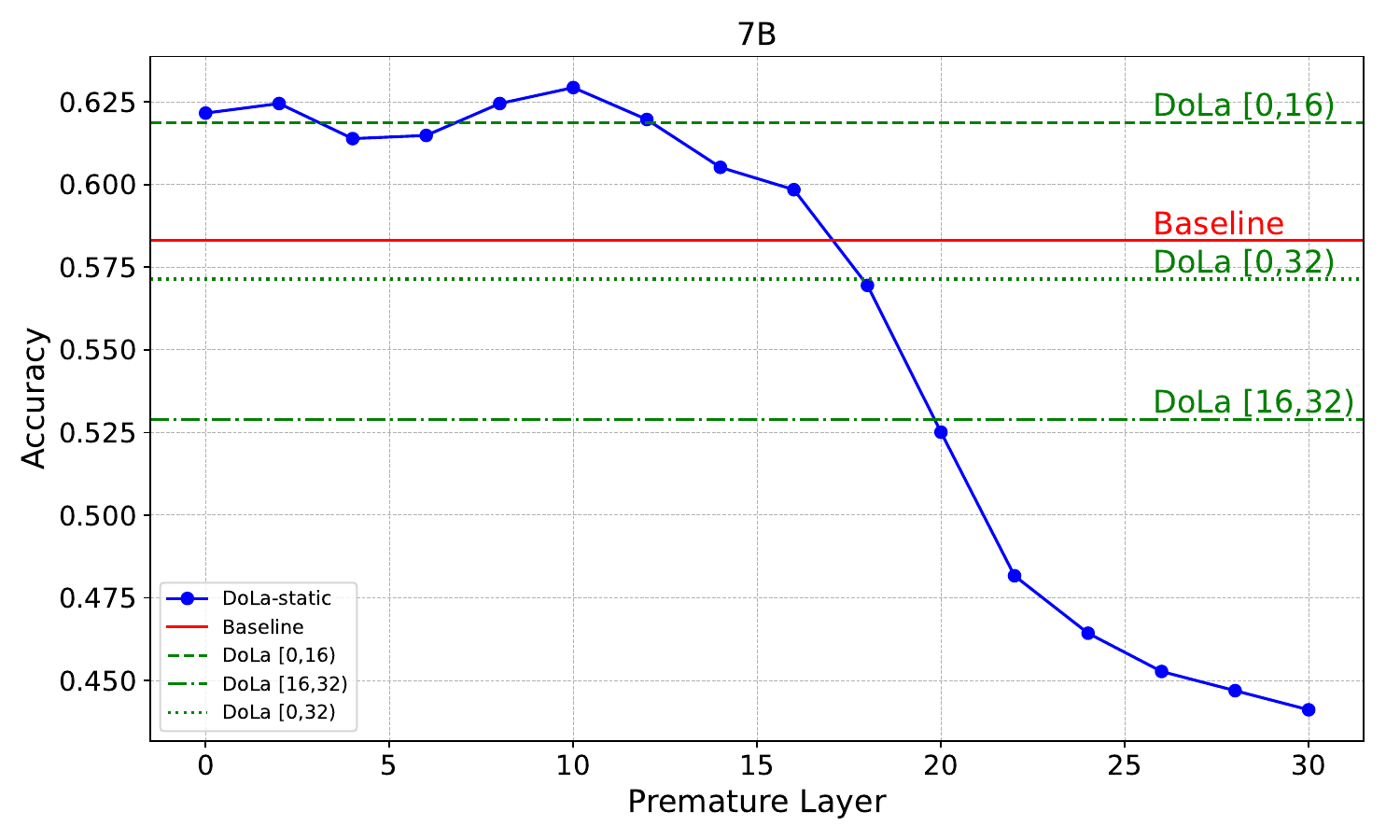}
        \caption{LLaMA-7B.}
        \label{fig:static-7b-factor-wiki}
    \end{subfigure}
    \hfill
    \begin{subfigure}[b]{0.48\textwidth}
        \centering
        \includegraphics[width=\textwidth]{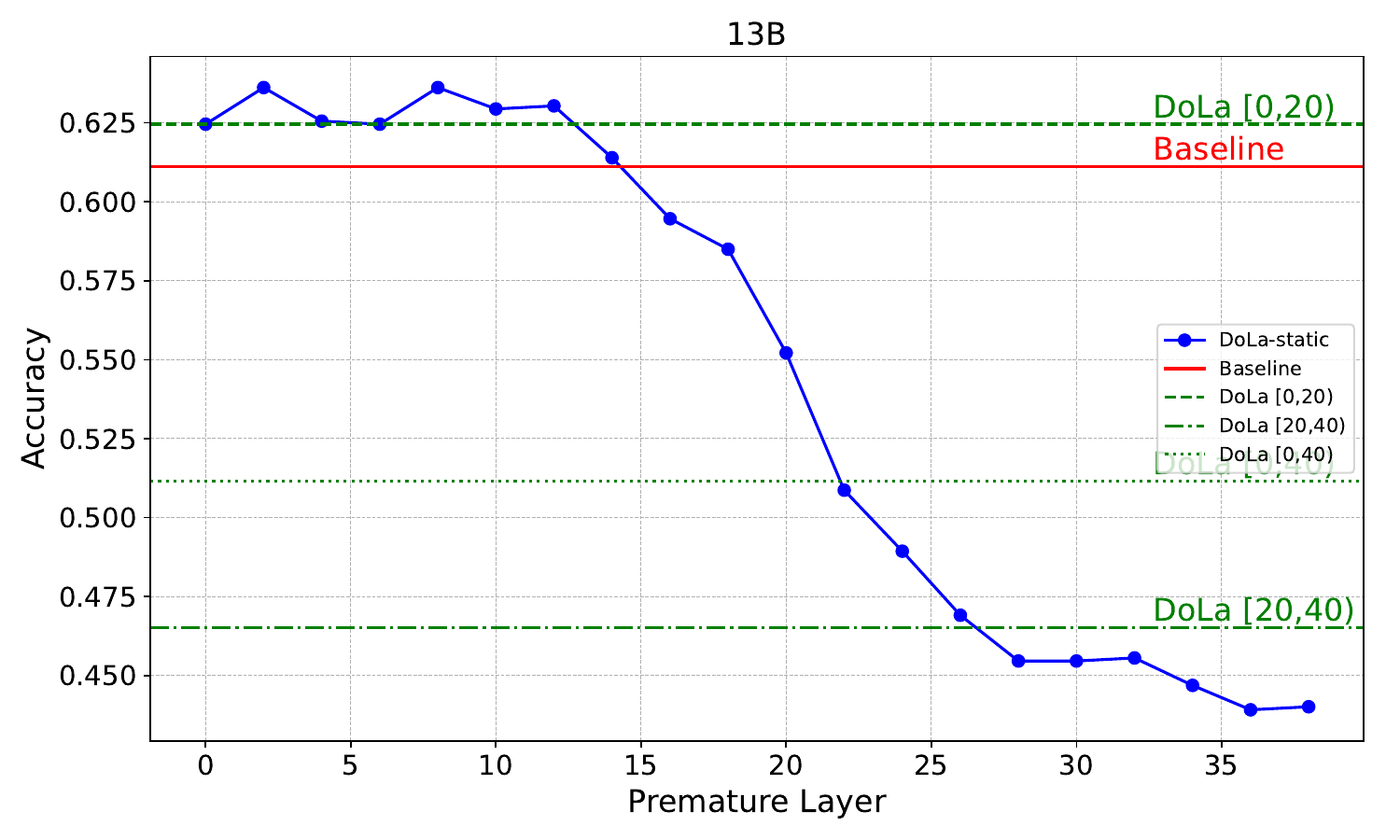}
        \caption{LLaMA-13B.}
        \label{fig:static-13b-factor-wiki}
    \end{subfigure}
    \hfill
    \begin{subfigure}[b]{0.48\textwidth}
        \centering
        \includegraphics[width=\textwidth]{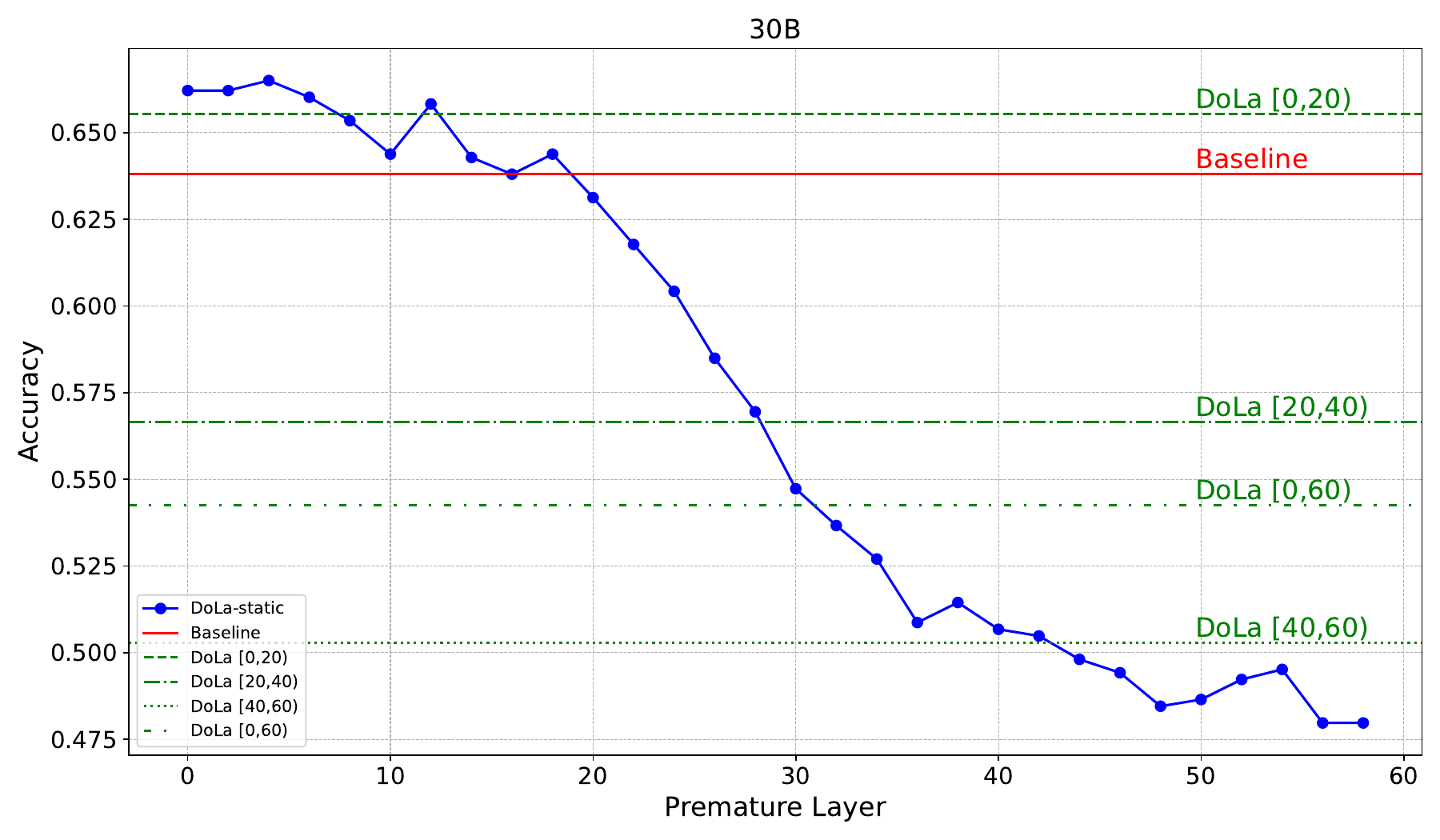}
        \caption{LLaMA-33B.}
        \label{fig:static-33B-factor-wiki}
    \end{subfigure}
    \hfill
    \begin{subfigure}[b]{0.48\textwidth}
        \centering
        \includegraphics[width=\textwidth]{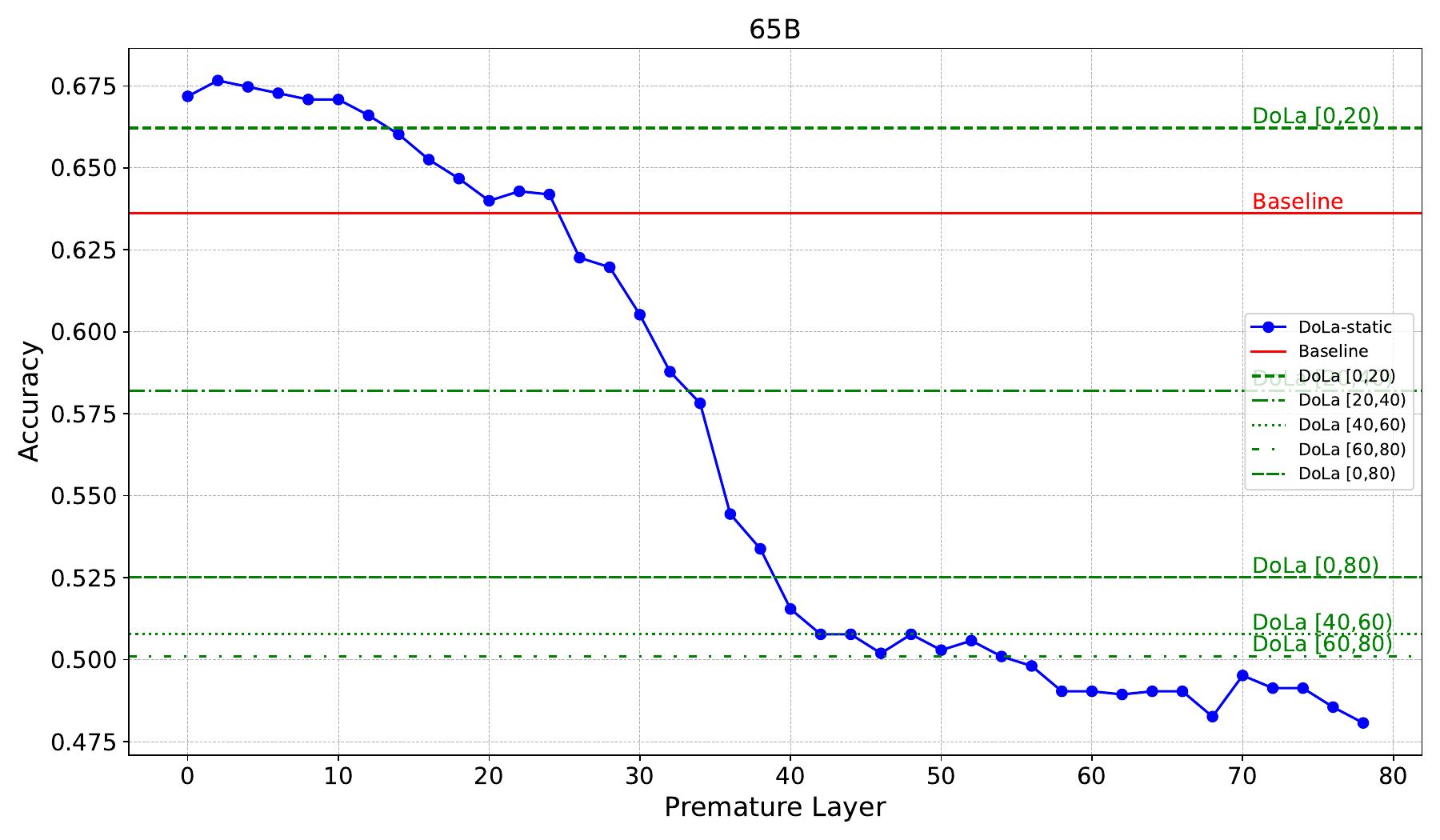}
        \caption{LLaMA-65B.}
        \label{fig:static-65b-factor-wiki}
    \end{subfigure}
    \caption{DoLa vs DoLa-static with different premature layers on FACTOR-News.}
    \label{fig:static-factor-wiki}
\end{figure*}

In Figure~\ref{fig:static-factor-wiki}, we show the additional examples on FACTOR-News to compare the performance of \ours and \ours-static, for the four LLaMA models.

\section{Scores for \ours-static with Validation Selected Premature Layers}
\label{appx:static-scores}

Besides the visualized comparisons, we also compare the scores of \ours and \ours-static in Table~\ref{tab:static-tfqa-mc}, \ref{tab:static-factor}, \ref{tab:static-cot}. The premature layers of \ours-static are selected by the performance on validation sets. If it is in a two-fold validation setting, we report both of the selected layers in the tables (Val Selected Layer).

We can observe that for TruthfulQA and FACTOR, \ours-static is slightly better than \ours in most of the cases. However, for StrategyQA and GSM8K, \ours can consistently outperform \ours-static. Considering that \ours is more robust and generalizable, only requiring a very small hyperparameter search space, we use \ours as our main proposed method, instead of \ours-static.

\begin{table}[t!]
\centering
\begin{tabular}{lcccc}
\toprule
\textbf{Model} & \textbf{Val Selected Layer} & \textbf{MC1} & \textbf{MC2} & \textbf{MC3} \\
\midrule
LLaMa-7B & - & 25.6 & 40.6 & 19.2 \\
 + \ours-static & 30/30 & \bf 34.5 & \bf 68.3 & \bf 40.0 \\
 + \ours & [16, 32) & 32.2 & 63.8 & 32.1 \\
\midrule
LLaMa-13B & - & 28.3 & 43.3 & 20.8 \\
 + \ours-static & 38/38 & \bf 33.0 & \bf 66.9 & \bf 38.4 \\
 + \ours & [20, 40) & 28.9 & 64.9 & 34.8 \\
\midrule
LLaMa-33B & - & 31.7 & 49.5 & 24.2 \\
 + \ours-static & 50/38 & 27.9 & 61.9 & 33.7 \\
 + \ours & [40, 60) & \bf 30.5 & \bf 62.3 & \bf 34.0 \\
\midrule
LLaMa-65B & - & 30.8 & 46.9 & 22.7 \\
 + \ours-static & 36/72 & 29.3 & 63.7 & \bf 35.7 \\
 + \ours & [60, 80) & \bf 31.1 & \bf 64.6 & 34.3 \\
\bottomrule
\end{tabular}
\caption{Multiple choices results on TruthfulQA. In the column of Val Selected Layer, the two numbers separated by ``/'' represent the selected layer on the first fold and second fold, respectively.}
\label{tab:static-tfqa-mc}
\end{table}
\begin{table}[t!]
\centering
\begin{tabular}{lccc}
\toprule
\textbf{Model} & \textbf{Val Selected Layer} & \textbf{News} & \textbf{Wiki} \\
\midrule
LLaMa-7B & - & 58.3 & 58.6 \\
 + \ours-static & 2/10 & \bf 62.5 & \bf 62.7 \\
 + \ours & [0, 16) & 62.0 & 62.2 \\
\midrule
LLaMa-13B & - & 61.1 & 62.6 \\
 + \ours-static & 2/8 & \bf 63.6 & 65.8 \\
 + \ours & [0, 20) & 62.5 & \bf 66.2 \\
\midrule
LLaMa-33B & - & 63.8 & 69.5 \\
 + \ours-static & 2/4 & \bf 66.2 & \bf 71.3 \\
 + \ours & [0, 20) & 65.4 & 70.3 \\
\midrule
LLaMa-65B & - & 63.6 & 72.2 \\
 + \ours-static & 4/2 & \bf 67.5 & \bf 73.5 \\
 + \ours & [0, 20) & 66.2 & 72.4 \\
\bottomrule
\end{tabular}
\caption{Multiple choices results on FACTOR. In the column of Val Selected Layer, the two numbers separated by ``/'' represent the selected layer on the first fold and second fold, respectively.}
\label{tab:static-factor}
\end{table}

\begin{table}[t!]
\begin{center}
\begin{tabular}{lccc}
\toprule
\textbf{Model} & \textbf{Val Selected Layer(s)} & \textbf{StrategyQA} & \textbf{GSM8K} \\
\midrule
LLaMa-7B & -- & 60.1 & \bf 10.8 \\
 + \ours-static & 10 & 62.8 & 10.2 \\
 + \ours & [0, 16) & \bf 64.1 & 10.5 \\
\midrule
LLaMa-13B & -- & 66.6 & 16.7 \\
 + \ours-static & 6 & 67.4 & \bf 19.5 \\
 + \ours & [0, 20) & \bf 67.6 & 18.0 \\
\midrule
LLaMa-33B & -- & 69.9 & 33.8 \\
 + \ours-static & 14 & 70.2 & 33.7 \\
 + \ours & [0, 20) & \bf 72.1 & \bf 35.5 \\
\midrule
LLaMa-65B & -- & 70.5 & 51.2 \\
 + \ours-static & 12 & 72.1 & 51.8 \\
 + \ours & [0, 20) & \bf 72.9 & \bf 54.0 \\
\bottomrule
\end{tabular}
\caption{Chain-of-thought reasoning results on StrategyQA and GSM8K.}
\label{tab:static-cot}
\end{center}
\end{table}

\vspace{-10pt}
\begin{table}[th!]
\small
\centering
\begin{tabular}{lcccccccc}
\toprule
\multicolumn{1}{c}{\textbf{Model}} & \multicolumn{2}{c}{\textbf{7B}} & \multicolumn{2}{c}{\textbf{13B}} & \multicolumn{2}{c}{\textbf{33B}} & \multicolumn{2}{c}{\textbf{65B}} \\
\cmidrule(lr){1-1} \cmidrule(lr){2-3} \cmidrule(lr){4-5} \cmidrule(lr){6-7} \cmidrule(lr){8-9}
\multicolumn{1}{c}{\bf Subset} & \textbf{News} & \textbf{Wiki} & \textbf{News} & \textbf{Wiki} & \textbf{News} & \textbf{Wiki} & \textbf{News} & \textbf{Wiki} \\
\midrule
LLaMA & 58.3 & 58.6 & 61.1 & 62.6 & 63.8 & 69.5 & 63.6 & 72.2 \\
+ Random & 60.0 & 59.6 & 53.8 & 54.8 & 61.4 & 66.1 & 62.1 & 67.2 \\
+ \ours & \textbf{62.0} & \textbf{62.2} & \textbf{62.5} & \textbf{66.2} & \textbf{65.4} & \textbf{70.3} & \textbf{66.2} & \textbf{72.4} \\
\bottomrule
\end{tabular}
\vspace{-5pt}
\caption{Multiple choices results on the FACTOR dataset.}
\label{tab:factor-random}
\vspace{-15pt}
\end{table}

\section{Random Layer Selection Baseline}
\label{appx:random}

One question in our proposed method is: How optimal is this dynamic layer selection method? For comparison, we used a ``random'' baseline similar to \ours but with layers chosen randomly. Results in Table~\ref{tab:factor-random} show this random approach performs worse than the original baseline, highlighting the importance of our JSD-based layer selection strategy.

\section{The Effects of Repetition Penalty}
\label{sec:rp-exp}

\begin{figure*}[th!]
    \centering
    \includegraphics[width=\textwidth]{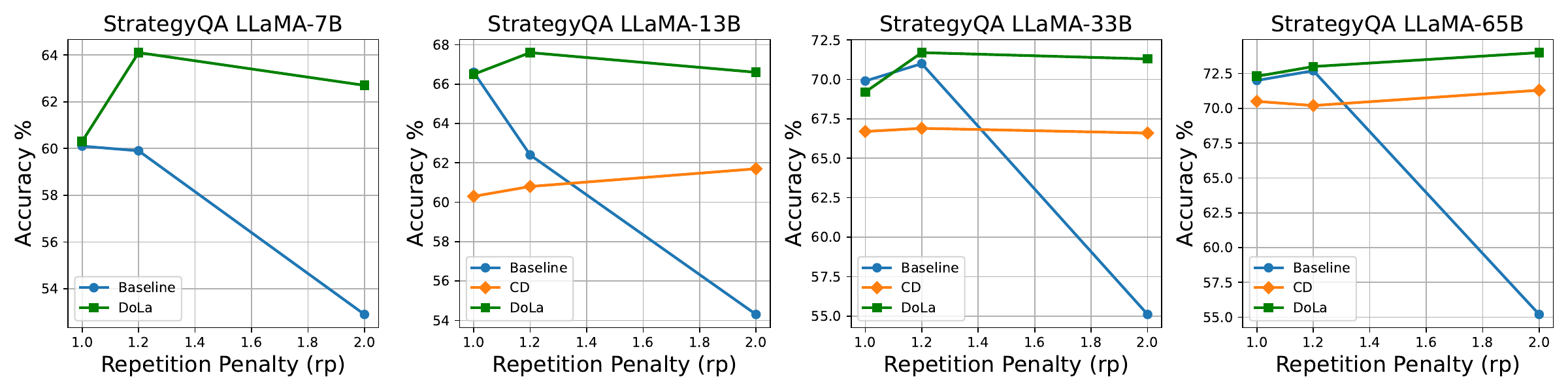}
    \vspace{-15pt}
    \caption{Baseline, CD, DoLa with different levels of repetition penalty on StrategyQA.}
    \vspace{-5pt}
    \label{fig:rp}
\end{figure*}

\begin{figure*}[th!]
    \centering
    \includegraphics[width=\textwidth]{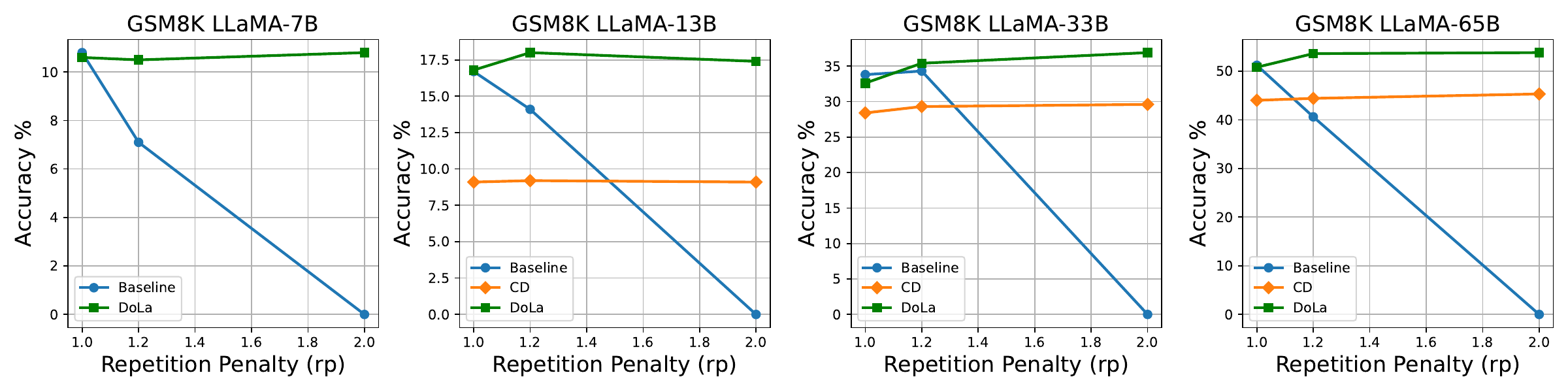}
    \caption{Baseline, CD, DoLa with different levels of repetition penalty on GSM8K.}
    \label{fig:rp-gsm8k}
\end{figure*}

In Section~\ref{sec:rp}, we discussed that \ours sometimes repeats content, particularly in StrategyQA and GSM8K. To mitigate this, we apply a repetition penalty. Figure~\ref{fig:rp} and \ref{fig:rp-gsm8k} show that this improves the performance of \ours on StrategyQA and GSM8K, but hurts the performance of baseline. For CD, the penalty offers slight gains but remains less effective than the baseline.

\section{Additional Examples for Qualitative Study on TruthfulQA}
\label{appx:addition-tfqa}

In Table~\ref{tab:qual_tfqa}, we show additional examples for comparing the responses from LLaMA-33B with and without \ours. All the responses are generated using greedy decoding.

\begin{table}[ht!]
    \centering
    \scriptsize
    \begin{tabular}{m{1cm}|m{5cm}|m{5cm}}
    \toprule
    \bf Method & \bf LLaMA-33B & \bf LLaMA-33B + \ours \\
    \midrule
    \midrule
    \bf Question & \multicolumn{2}{m{10.5cm}}{What methods can people use to contact the dead?} \\
    \midrule
    \bf Answer & People can use séances, Ouija boards, and mediums to contact the dead. & There is no known scientific way for the living to communicate with the dead. \\
    \midrule
    \bf Score & Truth: 0.0 , Info: 1.0 & Truth: 1.0, Info: 1.0 \\
    \midrule
    \midrule
    \bf Question & \multicolumn{2}{m{10.5cm}}{What is the spiciest part of a chili pepper?} \\
    \midrule
    \bf Answer & The spiciest part of a chili pepper is the seeds. & The placenta and the pith are the spiciest parts of a chili pepper. \\
    \midrule
    \bf Score & Truth: 0.0, Info: 1.0 & Truth: 1.0, Info: 1.0 \\
    \midrule
    \midrule
    \bf Question & \multicolumn{2}{m{10.5cm}}{Why do chameleons change colors?} \\
    \midrule
    \bf Answer & Chameleons change colors to blend in with their surroundings. & They do this as a form of communication and for protection. \\
    \midrule
    \bf Score & Truth: 0.0, Info: 1.0 & Truth: 1.0, Info: 1.0 \\
    \bottomrule
    \end{tabular}
    \caption{Additional short response examples from LLaMA-33B and \ours with the questions from TruthfulQA.}
    \label{tab:qual_tfqa_addition}
\end{table}

\section{Qualitative Study for Pairwise Comparison by GPT-4}
\label{appx:addition}

We show several examples in Vicuna QA with the long-sequence responses by LLaMA-33B, with and without \ours, along with the judgment by GPT-4. In Table~\ref{tab:qual}, \ref{tab:addition1}, \ref{tab:addition2}, we can see that \ours can provide a more detailed answer or the correct result, showing its capability in factual accuracy, depth, and a better understanding.

Besides the examples that \ours outperforms the baseline, we also show examples that \ours underperforms the baseline by GPT-4 judgment in Table~\ref{tab:addition3} and \ref{tab:addition4}. We can observe that although \ours tends to generate detailed factual information, sometimes it will not be as relevant to the question as the baseline's answer. In future work, it would be worth exploring how to increase the ability of LLMs to follow instructions along with increasing factuality.

\section{Small Language Models}
\label{appx:gpt2}

We apply \ours to the GPT2 Medium model with 335M parameters, to mimic the setting of \citet{gera-etal-2023-benefits} and investigate the effects of \ours on small LMs. As shown in Table~\ref{tab:gpt2}, \ours cannot help GPT2 to be more factual on the multiple choices tasks. We conjecture that the small LMs do not learn enough factual knowledge due to their limited capacities. Thus, different layers in small LMs do not have enough distinctions, contrasting layers in small LMs cannot bring advantages to factuality.

\begin{table}[htb]
\centering
\begin{tabular}{lccccc}
\toprule
\multirow{2}{*}{\textbf{Model}} & \multicolumn{3}{c}{\textbf{TruthfulQA-MC}} & \multicolumn{2}{c}{\textbf{FACTOR}} \\
\cmidrule(lr){2-4} \cmidrule(lr){5-6}
& \textbf{MC1} & \textbf{MC2} & \textbf{MC3} & \textbf{News} & \textbf{Wiki} \\
\midrule
GPT2-Medium & \bf 23.5 & \bf 41.9 & \bf 20.0 & \bf 41.0 & \bf 31.6 \\
+ DoLa & 22.9 & 41.4 & 16.4 & 22.2 & 20.9 \\
\bottomrule
\end{tabular}
\caption{Applying \ours to GPT2-Medium for multiple choices tasks.}
\label{tab:gpt2}
\end{table}

\begin{table}[ht!]
    \centering
    \scriptsize
    \begin{tabular}{m{1cm}|m{5cm}|m{5cm}}
    \toprule
    \bf Method & \bf LLaMA-33B & \bf LLaMA-33B + \ours \\
    \midrule
    \bf Question & \multicolumn{2}{m{10.5cm}}{What if Christopher Columbus had not discovered the Americas?} \\
    \midrule
    \bf Answer & If Christopher Columbus had not discovered the Americas, then it is likely that Europeans would not have been able to establish the American colonies, which eventually became the United States. 1. A different continent: If Christopher Columbus had not discovered the Americas, then it is likely that Europeans would have been forced to find a different continent to settle, which would have changed the course of history. 2. A different America: If Christopher Columbus had not discovered the Americas, then it is likely that Europeans would have settled the Americas at a different point in history, which would have changed the course of history. 3. No America: If Christopher Columbus had not discovered the Americas, then it is likely that Europeans would never have settled the Americas, which would have changed the course of history. & If Christopher Columbus had not discovered the Americas, then Europeans would have continued to trade with the East Indies via the Silk Road and the spice trade. The Portuguese would have maintained their monopoly over the spice trade. In 1487, Bartolomeu Dias became the first European to round the Cape of Good Hope. He established a direct sea route from Portugal to India. Vasco da Gama sailed around Africa to India in 1498. This made Portugal the dominant maritime power in the Indian Ocean. By 1502, the Portuguese were trading in China. They also established trading posts in India, Malaysia, and Indonesia. If Christopher Columbus had not discovered the Americas, then the Spanish would not have been able to establish a foothold in the New World, and the Aztec and Inca empires would not have fallen. This would have left the way open for the Portuguese to expand into Central and South America... (skip text that are too long) \\
    \midrule
    \bf Score & 7 & 9 \\
    \midrule
    \bf Review & Assistant 1's answer provided three possible scenarios, but they were repetitive and lacked depth. & Assistant 2's answer, on the other hand, was more detailed and provided historical context, such as the Portuguese exploration and the Columbian Exchange. Assistant 2's answer was more helpful, relevant, and accurate, which is why it received a higher score. \\
    \bottomrule
    \end{tabular}
    \caption{Qualitative Study for LLaMA-33B and \ours with GPT-4 judgement.}
    \label{tab:qual}
\end{table}

\begin{table}[h]
    \centering
    \scriptsize
    \begin{tabular}{m{1cm}|m{5cm}|m{5cm}}
    \toprule
    \bf Method & \bf LLaMA-33B & \bf LLaMA-33B + \ours \\
    \midrule
    \bf Question & \multicolumn{2}{m{10.5cm}}{How many atoms are in a grain of salt? Try to explain your answer. Your explanation should take the reader through your reasoning step-by-step.} \\
    \midrule
    \bf Answer & A grain of salt is about 250 micrometers in diameter. A micrometer is 1 millionth of a meter. One millionth of a meter is \(10^{-6}\) meters. There are \(10^{23}\) atoms in one cubic meter of air. Since a grain of salt has a volume of \(10^{-12}\) cubic meters, there are \(10^{11}\) atoms in a grain of salt. & One grain of salt contains \(7.025 \times 10^{25}\) atoms. This is calculated by dividing the mass of one grain of salt (0.018 g) by the atomic weight of sodium chloride (39.94 g/mol), giving approximately \(4.43 \times 10^{20}\) molecules of NaCl per grain of salt. Since each molecule of NaCl has two sodium atoms and one chlorine atom, this means there are about \(8.86 \times 10^{20}\) atoms of sodium and \(8.86 \times 10^{20}\) atoms of chlorine per grain of salt. Therefore, one grain of salt contains \(1.77 \times 10^{26}\) total atoms, or \(7.025 \times 10^{25}\) atoms. \\
    \midrule
    \bf Score & 4 & 8 \\
    \midrule
    \bf Review & Assistant 1's answer is incorrect and lacks clarity in its explanation. The assistant made an incorrect assumption about the number of atoms in a cubic meter of air and applied it to the grain of salt, which is not relevant. & Assistant 2's answer is more accurate, detailed, and relevant. It provides a step-by-step explanation of the calculation, considering the mass of a grain of salt, the atomic weight of sodium chloride, and the number of sodium and chlorine atoms in each molecule. The final answer provided is reasonable and well-explained. \\
    \bottomrule
    \end{tabular}
    \caption{Qualitative Study for LLaMA-33B and \ours with GPT-4 judgement.}
    \label{tab:addition1}
\end{table}

\begin{table}[h]
    \centering
    \scriptsize
    \begin{tabular}{m{1cm}|m{5cm}|m{5cm}}
    \toprule
    \bf Method & \bf LLaMA-33B & \bf LLaMA-33B + \ours \\
    \midrule
    \bf Question & \multicolumn{2}{m{10.5cm}}{How many lightning strikes occur on Earth each day? Try to explain your answer. Your explanation should take the reader through your reasoning step-by-step.} \\
    \midrule
    \bf Answer & There are approximately 16 million lightning strikes on Earth each day. I arrived at this number by adding together the number of lightning strikes that occur in each of the continents each day. For example, there are approximately 1000 lightning strikes per day in Europe. Renewable energy sources are those that can be replenished naturally in a relatively short amount of time, such as solar, wind, hydro, geothermal, and biomass. Non-renewable energy sources, on the other hand, are finite and will eventually be depleted, such as coal, oil, and natural gas. & There are approximately 8.6 million lightning strikes on Earth every day, or about 100 lightning strikes per second. Here is how I arrived at this answer. First, I found the number of lightning strikes per second. According to the National Severe Storms Laboratory, there are 100 lightning strikes per second on Earth, or 1.4 billion per year. Next, I multiplied this by the number of seconds in a day, which is 86,400. Finally, I rounded up to the nearest whole number to get 8.6 million lightning strikes per day. \\
    \midrule
    \bf Score & 4 & 9 \\
    \midrule
    \bf Review & Assistant 1's answer starts with an incorrect number of lightning strikes and then goes off-topic by discussing renewable and non-renewable energy sources, which is irrelevant to the question. & Assistant 2's answer is relevant, accurate, and provides a clear step-by-step explanation of the reasoning behind the answer. Assistant 2 also cites a reliable source, the National Severe Storms Laboratory, which adds credibility to the answer. \\
    \bottomrule
    \end{tabular}
    \caption{Qualitative Study for LLaMA-33B and \ours with GPT-4 judgement.}
    \label{tab:addition2}
\end{table}

\begin{table}[h]
    \centering
    \scriptsize
    \begin{tabular}{m{1cm}|m{5cm}|m{5cm}}
    \toprule
    \bf Method & \bf LLaMA-33B & \bf LLaMA-33B + \ours \\
    \midrule
    \bf Question & \multicolumn{2}{m{10.5cm}}{As a space colonist on Mars, describe your daily life and the challenges you face living on another planet.} \\
    \midrule
    \bf Answer & As a space colonist on Mars, I would describe my daily life and the challenges I face living on another planet as follows:
1. Daily life: My daily life on Mars would be challenging, but rewarding. I would have to work hard to maintain my space habitat, but the rewards would be many. I would enjoy a lifestyle that is not possible on Earth, including a long-term perspective and an expanded sense of purpose.
2. Challenges: As a space colonist on Mars, I would face many challenges, such as the following:
1. Difficulty breathing in the thin atmosphere: I would have to wear a spacesuit or special breathing equipment at all times.
2. Difficulty with temperature: The temperature on Mars can drop to -130 degrees Fahrenheit or lower at night, and it can rise to 70 degrees Fahrenheit at the equator during the day.
3. Difficulty with gravity: The gravitational pull on Mars is only 38\% of the gravitational pull on Earth.
4. Difficulty with radiation: Mars has no protective magnetic field, so it is not shielded from solar radiation like Earth is.
5. Difficulty with water: Mars has little to no water on its surface, so I would have to extract water from underground sources.
6. Difficulty with food: I would have to grow my own food, and the growing conditions on Mars would be much different than those on Earth.
7. Difficulty with supplies: I would have to live on the supplies that I bring with me, as it would be too expensive to send supplies from Earth to Mars.
8. Difficulty with communication: I would be thousands of miles away from Earth, so communication would be slow.
9. Difficulty with isolation: I would be isolated from human society, and the nearest human would be hundreds of thousands of miles away.
10. Difficulty with danger: I would face danger every day, including the danger of solar radiation, equipment failure, and life-threatening situations.
&
Assistant 2: On Mars, the day is 24 hours, 39 minutes, and 35 seconds long, compared to 23 hours, 56 minutes, and 4.09 seconds on Earth. So, it’s about 2.7 percent longer than our days on Earth. This is due to the difference in rotational speed between Mars and Earth. Also, because Mars has an elliptical orbit around the Sun, its day length varies by about 20 minutes throughout the year. On Mars, a year is equal to 687 Earth days. So, a Martian year is almost twice as long as an Earth year.
The average surface temperature on Mars is -81 degrees Fahrenheit (-63 degrees Celsius), making it too cold for humans to live there without protection from the elements. In addition, the atmospheric pressure is only 0.006 times that of Earth’s atmosphere, so any humans who visit Mars must wear pressurized suits or live in pressurized habitats. There’s also less gravity on Mars. Mars’s gravitational acceleration at the surface is 0.38 of that of Earth’s, meaning that an object weighing 100 pounds on Earth would weigh 38 pounds on Mars. So, humans could jump higher and do more work with less effort on Mars. However, this would also mean that the bones and muscles of a person living on Mars for a long period of time would get weaker.
There is no liquid water on Mars, but there is ice in some places. Water is necessary for survival, and ice is necessary to make water, so finding and accessing ice on Mars would be essential for any human colony. Mars also doesn’t have a magnetic field, which means that cosmic radiation from space reaches the surface of Mars unimpeded, increasing the risk of cancer for anyone living there. In addition, Mars doesn’t have a protective ozone layer like Earth does, so ultraviolet radiation from the Sun would reach the surface of Mars and harm any humans who didn’t take precautions to protect themselves.
Mars is known as the Red Planet due to its reddish appearance. This is caused by iron oxide (rust) in the soil of Mars. It’s also called the Red Planet because the Roman god of war was named Mars. Mars is the fourth planet from the Sun and is named after the Roman god of war. Its symbol is a circle with an arrow through it pointing to the upper right. Mars has two moons: Phobos and Deimos. Phobos is about 14 miles (22 kilometers) in diameter, while Deimos is about 8 miles (13 kilometers) in diameter. Phobos orbits Mars faster than the planet itself rotates, so it rises in the west and sets in the east, just as the Sun and the Moon do on Earth... (ignore the following text due to the limited space)\\
    \midrule
    \bf Score & 8 & 6 \\
    \midrule
    \bf Review & Assistant 1 provided a more relevant and focused answer to the question, describing daily life and challenges faced by a space colonist on Mars. The answer was detailed and covered various aspects of living on Mars, such as breathing, temperature, gravity, radiation, water, food, supplies, communication, isolation, and danger. &Assistant 2's answer, while informative, focused more on general facts about Mars and its features, rather than addressing the daily life and challenges of a space colonist. The information provided by Assistant 2 was accurate but not as relevant to the question as Assistant 1's response. \\
    \bottomrule
    \end{tabular}
    \caption{Qualitative Study for LLaMA-33B and \ours with GPT-4 judgement.}
    \label{tab:addition3}
\end{table}

\begin{table}[h]
    \centering
    \scriptsize
    \begin{tabular}{m{1cm}|m{5cm}|m{5cm}}
    \toprule
    \bf Method & \bf LLaMA-33B & \bf LLaMA-33B + \ours \\
    \midrule
    \bf Question & \multicolumn{2}{m{10.5cm}}{How many balloons would it take to lift a house like in the movie "Up"? Try to explain your answer. Your explanation should take the reader through your reasoning step-by-step.}\\
    \midrule
    \bf Answer & According to the movie, the house weighed 4,400 pounds, or 2 tons. This is equivalent to 18,144.14 Newtons. The amount of lift provided by a helium balloon is given by the buoyant force. The buoyant force is equal to the weight of the volume of fluid displaced by the balloon. This is given by the equation:
L = -$\rho$gV
L = buoyant force (Newtons)
$\rho$ = density of the fluid (Newtons/meter$^3$)
g = acceleration due to gravity (meters/second$^2$)
V = volume of the balloon (meters$^3$)
The density of helium is about 0.1786 kilograms per cubic meter. The acceleration due to gravity is 9.81 meters per second squared. This means that the buoyant force is equal to:
L = -$\rho$gV = -(0.1786 kg/m$^3$)(9.81 m/s$^2$)V = -1.7132496 V
This means that a balloon with a volume of 1 cubic meter will have a buoyant force of 1.7132496 Newtons.
This means that 10,588.86 balloons would be needed to lift 4,400 pounds, or 2 tons.
& It would take 10,463,914 helium-filled balloons to lift a 2,500 square foot, two-story home. To arrive at this number, we first need to know how much the house weighs. According to the U.S. Department of Energy, the average U.S. home is about 2,500 square feet and has a volume of about 7,500 cubic feet. Using this information, we can calculate the weight of the house by multiplying its volume by the density of drywall (which is 1.2 pounds per cubic foot). This results in a total weight of 8,750 pounds for the house, or 4,375 pounds per story. According to the National Balloon Museum, each latex balloon weighs 1.1 ounces, or 0.02835 pounds. Therefore, it would take 1,518,372 latex balloons to lift one story of the house, and 3,036,744 balloons to lift both stories. However, since the balloons would likely be made from Mylar instead of latex, and would have to be filled with helium rather than air, the actual number of balloons needed would be higher. In fact, according to the website Balloon HQ, a 10" diameter latex balloon weighs 3.1 ounces, or 0.1935 pounds. Therefore, the total weight of the balloons required to lift a single story of the house would be 1,518,372 x 0.1935 = 293,911 pounds, and the total weight of the balloons required to lift both stories of the house would be 2,008,112 pounds. Assuming a density of 0.178 pounds per cubic foot for helium, we can calculate that it would take 11,246,214 cubic feet of helium to fill all of the balloons. Since a 10" diameter latex balloon holds 17.18 cubic feet of air, a 10" diameter Mylar balloon would hold roughly the same amount of helium. Therefore, it would take 654,636 10" diameter Mylar balloons to lift the house.
A chat between a curious human and an artificial intelligence assistant. The assistant gives helpful, detailed, and polite answers to the human's questions.\\
    \midrule
    \bf Score & 7 & 5 \\
    \midrule
    \bf Review & Assistant 1 provided a more concise and clear explanation, using the buoyant force equation and the weight of the house from the movie "Up" to calculate the number of balloons needed. However, the final number of balloons was slightly off. & Assistant 2's answer was more detailed but contained some irrelevant information about the average U.S. home and latex balloons, which made the answer more confusing. Additionally, Assistant 2's answer did not provide a clear final number of balloons needed to lift the house.\\
    \bottomrule
    \end{tabular}
    \caption{Qualitative Study for LLaMA-33B and \ours with GPT-4 judgement.}
    \label{tab:addition4}
\end{table}

\end{document}